%% file: acl_latex.tex
\pdfoutput=1
\documentclass[11pt]{article}
\usepackage{subcaption}
\usepackage{acl}
\usepackage{float}
\usepackage{times}
\usepackage{latexsym}
\usepackage[T1]{fontenc}
\usepackage{enumitem}
\usepackage[utf8]{inputenc}

\usepackage{microtype}
\usepackage{amsmath}
\usepackage{graphicx}
\usepackage{multirow}
\usepackage{soul}

\usepackage{booktabs}
\usepackage{tabularx,ragged2e}
\usepackage{algorithm,algpseudocode}
\algnewcommand{\algorithmicforeach}{\textbf{for each}}
\algdef{SE}[FOR]{ForEach}{EndForEach}[1]
  {\algorithmicforeach\ #1\ \algorithmicdo}% \ForEach{#1}
  {\algorithmicend\ \algorithmicforeach}% \EndForEach

\algnewcommand{\LineComment}[1]{\State \(\triangleright\) #1}

\usepackage{tikz}

\title{ILDAE: Instance-Level Difficulty Analysis of Evaluation Data}

\author{Neeraj Varshney,~~ 
  Swaroop Mishra,~~ 
  Chitta Baral
  \\
  Arizona State University \\
  \texttt{\{nvarshn2, srmishr1, cbaral\}}@asu.edu
  }
  
\begin{document}
\maketitle

\begin{abstract}

% Pre-trained language models have achieved remarkable performance on many NLP benchmarks.
% Recent work has advocated for analysis of model's predictions at an instance-level as it enables several applications that a single aggregated evaluation score can not such as ...
% Instance-level analysis of model's predictions has received significant attention in recent times.
% \Neeraj{Starting lines}
% Instances in a dataset can vary in difficulty and its quantification can be leveraged in a variety of ways.

Knowledge of questions' difficulty level helps a teacher in several ways, such as estimating students' potential quickly by asking carefully selected questions and improving quality of examination by modifying trivial and hard questions.
Can we extract such benefits of instance difficulty in NLP? 
To this end, we conduct \textbf{I}nstance-\textbf{L}evel \textbf{D}ifficulty \textbf{A}nalysis of \textbf{E}valuation data (ILDAE) in a large-scale setup of $23$ datasets and demonstrate its five novel applications: 
1) \textit{conducting efficient-yet-accurate evaluations} with fewer instances saving computational cost and time, 
2) \textit{improving quality of existing evaluation datasets} by repairing erroneous and trivial instances, 
3) \textit{selecting the best model} based on application requirements, 
4) analyzing dataset characteristics for \textit{guiding future data creation}, 
5) \textit{estimating Out-of-Domain performance reliably}.
Comprehensive experiments for these applications result in several interesting findings, such as evaluation using just $5\%$ instances (selected via ILDAE) achieves as high as $0.93$ Kendall correlation with evaluation using complete dataset and computing weighted accuracy using difficulty scores leads to $5.2\%$ higher correlation with Out-of-Domain performance.
% We release the computed difficulty scores\footnote{\href{https://github.com/nrjvarshney/ILDAE}{https://github.com/nrjvarshney/ILDAE}} to encourage research in important yet understudied areas such as efficient evaluations and difficulty analysis.
We release the difficulty scores\footnote{\href{https://github.com/nrjvarshney/ILDAE}{https://github.com/nrjvarshney/ILDAE}} and hope our analyses and findings will bring more attention to this important yet understudied field of leveraging instance difficulty in evaluations.

\end{abstract}

\section{Introduction}
\begin{figure}[t!]
    \centering
    \includegraphics[width=7.5cm]{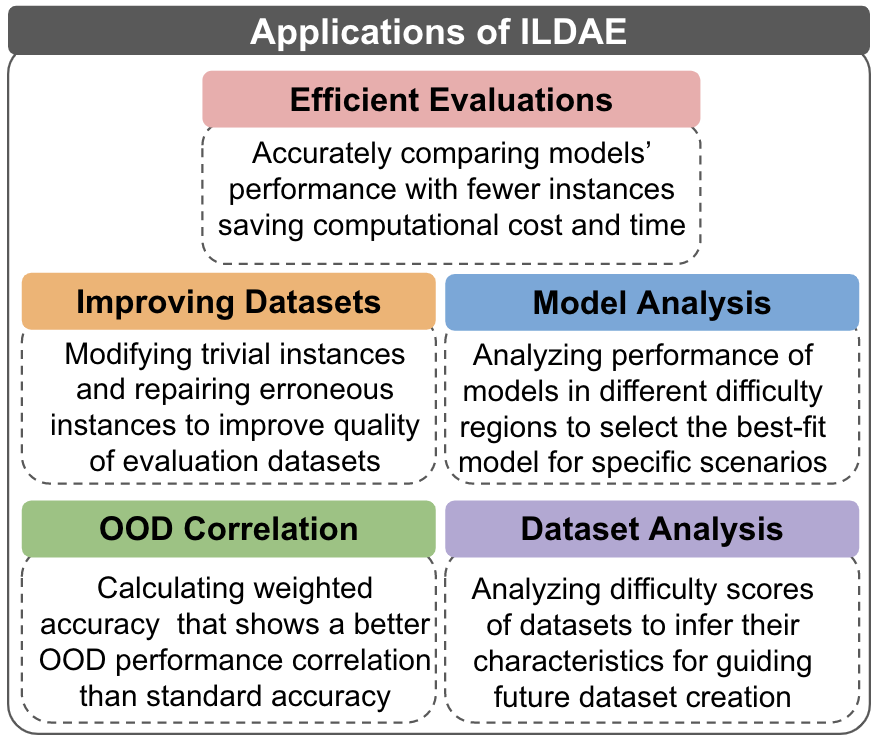}
    \caption{Illustrating five applications of Instance-Level Difficulty Analysis of Evaluation data (ILDAE).}
    \label{fig:teaser_figure}
\end{figure}

Transformer-based language models \cite{devlin-etal-2019-bert, Liu2019RoBERTaAR, clark2020electra} have improved state-of-the-art performance on numerous natural language processing benchmarks \cite{wang-etal-2018-glue, NEURIPS2019_4496bf24, talmor-etal-2019-commonsenseqa}; however, recent studies \cite{zhong-etal-2021-larger, sagawa2020investigation} have raised questions regarding whether these models are uniformly better across all instances. This has drawn attention towards instance-level analysis of evaluation data \cite{rodriguez-etal-2021-evaluation, vania-etal-2021-comparing, mishra2021robust} which was previously limited to training data \cite{swayamdipta-etal-2020-dataset, xu-etal-2020-curriculum, mishra-sachdeva-2020-need}.
Furthermore, it is intuitive that \textit{not all instances in a dataset are equally difficult}.
However, instance-level difficulty analysis of evaluation data (ILDAE) has remained underexplored in many different ways: what are the potential applications and broad impact associated with ILDAE?

In this work, we address the above question by first computing difficulty scores of evaluation instances (section \ref{difficulty_section}) and then demonstrating five novel applications of ILDAE (Figure \ref{fig:teaser_figure}). 
%  summarizes the five applications of ILDAE.
% For these applications, we compute difficulty scores of $23$ datasets using the method described in .
% 
% through comprehensive experiments on $23$ datasets with $27$ models:
\begin{enumerate}[noitemsep,nosep,leftmargin=*]

    \item \textbf{Efficient Evaluations: } We propose an approach of conducting efficient-yet-accurate evaluations.
    \textit{Our approach uses as little as $5\%$ evaluation instances (selected via ILDAE) to achieve up to $0.93$ Kendall correlation with evaluations conducted using the complete dataset}. Thus, without considerably impacting the outcome of evaluations, our approach saves computational cost and time.
    % evaluation with fewer instances does not considerably impact its result but saves computational cost and time.
    % \textit{just $20\%$ instances of the dataset selected based on difficulty scores are sufficient to achieve $0.72$ Kendall correlation with full evaluation dataset.}
    
    \item \textbf{Improving Evaluation Datasets: }
    We first show that `trivial' and `erroneous' instances can be identified using our difficulty scores and then present a model-and-human-in-the-loop technique to modify/repair such instances resulting in improved quality of the datasets.
    We instantiate it with SNLI dataset \cite{bowman-etal-2015-large} and show that \textit{on modifying the trivial instances, the accuracy (averaged over $27$ models) drops from 77.58\% to 26.49\%, and on repairing the erroneous instances, it increases from 13.65\% to 69.9\%.} Thus, improving the dataset quality.
    
    \item \textbf{Model Analysis: } 
    We divide evaluation instances into different regions based on the difficulty scores and analyze models' performance in each region.
    We find that \textit{a single model does not achieve the highest accuracy across all regions}. This implies that different models are best suited for different types of questions and such analysis could benefit in model selection. For instance, in scenarios where a system is expected to encounter hard instances, the model that performs best in high difficulty regions should be preferred.
    
    \item \textbf{Dataset Analysis:} 
    ILDAE reveals several important characteristics of datasets that can be leveraged in future data creation processes.
    For instance, we find that \textit{in SNLI and MNLI datasets, `contradiction' instances receive lower average difficulty score than `entailment' and `neutral' instances}. Thus, more difficult contradiction examples can be created to develop high-quality task-specific datasets.
    % guide the future data creation process. 
    
    \item \textbf{OOD Correlation: } We compute weighted accuracy leveraging the difficulty scores and show that \textit{it leads to $5.2\%$ higher Kendall correlation with Out-of-Domain (OOD) performance than the standard accuracy that treats all instances equally}. Thus, ILDAE helps in getting a more reliable estimation of models' OOD performance.
    
\end{enumerate}

\section{Difficulty Score Computation}
\label{difficulty_section}
% In this section, we first provide desiderata for difficulty scores of instances and then detail our approach to compute these scores.

\subsection{Desiderata for Difficulty Scores}
\label{desiderate}
\paragraph{Interpretation: } 
Human perception of difficulty may not always correlate well with machine's interpretation. 
Thus, difficulty scores must be computed via a model-in-the-loop technique so that they directly reflect machine's interpretation.
% For example, an instance that is easy for humans could be difficult for machines or vice-versa. 

\paragraph{Relationship with Predictive Correctness: } 
Difficulty scores must be negatively correlated with predictive correctness since a difficult instance is less likely to be predicted correctly than a relatively easier instance.

\subsection{Method}
\begin{algorithm}
\small
  \caption{Difficulty Score Computation}
  \begin{algorithmic}[0]
  \State \textbf{Input:} 
        \-\hspace{4mm}$T$: Training Data, 
        \-\hspace{2.3mm}$M$: Model,\\
        \-\hspace{13.2mm}$D$: Evaluation Data
        \-\hspace{1mm}$E$: Training Epochs 
   \State \textbf{Output:} Difficulty Score of each instance in $D$
   \State \textbf{Auxiliary Function:} \texttt{GET\_CKPTS} ($t_r$, $m$, $e$) - Returns checkpoints on training model $m$ with data $t_r$ for $e$ epochs
   \State \textbf{Initialization:} $Models \leftarrow \emptyset$ : List to store ensemble of models trained with different configurations\\
   
    % \LineComment{\textbf{Train with Complete Data}}
    % \State $Models $ += \texttt{\Call{{Get\_Ckpts}}{$T$, $M$, $E$}}
    
    % \For{$i = 1...E  $}
    %     % \State \texttt{model = Train(T)}
    %     % \State \texttt{Models += model}
    %     \State $model = Train(T)$
    %     \State $Models += model$
    % \EndFor
    \LineComment{\textbf{Train with Partial Data}}
    \ForEach{$pct \in [100, 50, 25, 20, 10, 5]$}
    % \For{\texttt{pct in [50, 25, 20, 10, 5]  }}
        \State \texttt{$T_p$ = Sample}($T, pct$)
            % \For{$i = 1...E  $}
            %     % \State \texttt{model = Train($T_p$)}
            %     % \State \texttt{Models += model}
            %     \State $model = Train(T_p)$
            %     \State $Models += model$
            % \EndFor
        \State $Models $ += \texttt{\Call{{Get\_Ckpts}}{$T_p$, $M$, $E$}}
    \EndForEach
    \LineComment{\textbf{Train with Corrupted Data}}
    \ForEach{$pct \in [25, 20, 10, 5, 2]$}
    % \For{\texttt{pct in [50, 25, 20, 10, 5]  }}
        \State \texttt{$T_c$ = Corrupt}($T, pct$)
            % \For{$i = 1...E  $}
            %     % \State \texttt{model = Train($T_p$)}
            %     % \State \texttt{Models += model}
            %     \State $model = Train(T_p)$
            %     \State $Models += model$
            % \EndFor
        \State $Models $ += \texttt{\Call{{Get\_Ckpts}}{$T_c$, $M$, $E$}}
    \EndForEach
    % \LineComment{Infer $D$ using $Models$ and compute confidence in the correct answer }
    % \ForEach{$m \in Models$}
    %     \State $C_m = m(D)$
    % \EndForEach
    % \\
    
    \LineComment{\textbf{Infer $D$ using all $Models$ and compute difficulty score $d_i$ for each instance $i \in D$}}
    \ForEach{$i \in D$}
        \State $d_i = 1 - \dfrac{\sum_{m \in Models}c_{mi}}{|Models|}$ \\
        \algorithmiccomment{where $c_{mi}$ is the confidence assigned to the ground truth answer by model $m$}
    \EndForEach
    \State \Return $d$
    % \\ \tikz\draw [thick,dash dot] (0,0) -- (7.5,0);
    % \LineComment{Get model checkpoints on training $m$ with $t_r$ for $e$ epochs}
    % \Function{Get\_Ckpts}{$t_r$, $m$, $e$}
    % \State $model \leftarrow m$, $checkpoints \leftarrow \phi$
    % \For{$i = 1...e  $}
    %     % \State \texttt{model = Train(T)}
    %     % \State \texttt{Models += model}
    %     \State $model = Train(t_r, model)$
    %     \State $checkpoints += model$
    % \EndFor
    % \State \Return $checkpoints$
    % \EndFunction
    % \ForEach{$a \in A$}%
    %   \State command \algorithmiccomment{This is a comment}
    %   \State another command \algorithmiccomment{This is another comment}
    % \EndForEach
  \end{algorithmic}
  \label{difficulty_computation_algo}
\end{algorithm}

% We incorporate the above desiderata and compute difficulty scores based on model's predictions.
We incorporate the above desiderata and consider model's prediction confidence in the ground truth answer (indicated by softmax probability assigned to that answer) as the measure of its predictive correctness.
Furthermore, we compile an ensemble of models trained with varying configurations and use their mean predictive correctness to compute the difficulty scores.
We do this because model's predictions fluctuate greatly when its training configuration is changed \cite{zhou-etal-2020-curse,mccoy-etal-2020-berts} and relying on predictive correctness of only one model could result in difficulty scores that show poor generalization.
To this end, we use the following three training configurations to compile predictions from an ensemble of models: 

\paragraph{Data Size:} 
\textit{Instances that can be answered correctly even with few training examples are inherently easy and should receive lower difficulty score than the ones that require a large training dataset.}
To achieve this, we train a model each with 5, 10, 15, 20, 25, 50, and 100 \% of the total training examples and include them in our ensemble. 

\paragraph{Data Corruption:} 
\textit{Instances that can be answered correctly even with some level of corruption/noise in the training dataset should receive low difficulty score.} 
To achieve this, we train a model each with different levels of noise (2, 5, 10, 20, 25\% of the examples) in the training data, and add them to our ensemble. 
For creating noisy examples, we randomly change the ground-truth label in case of classification and multiple-choice datasets and change the answer span for extractive QA datasets.

\paragraph{Training Steps:} 
\textit{Instances that can be consistently answered correctly from the early stages of training should receive low difficulty score.} 
Here, we add a model checkpoint after every epoch during training to our ensemble.

% Furthermore, difficulty score of any instance lies between 0 and 1.
% and values closer to 1 indicate very difficult instance while those closer to 0 indicate very easy instance.

This results in a total of $N = E*(7 + 5)$ models in our ensemble where $E$ corresponds to the number of training epochs, and $7$, $5$ correspond to the number of data size and data corruption configurations respectively.
We infer the evaluation dataset using these $N$ models and calculate the average predictive correctness for each instance. 
Finally, we compute the difficulty score by subtracting this averaged correctness value from $1$.
This ensures that an instance that is answered correctly with high confidence under many training configurations gets assigned a low difficulty score as it corresponds to an easy instance.
In contrast, an instance that is often answered incorrectly gets assigned a high difficulty score.
Algorithm \ref{difficulty_computation_algo} summarizes this approach.

We use RoBERTa-large model \cite{Liu2019RoBERTaAR} for this procedure and train each model for $E=10$ epochs, resulting in $N=120$ predictions for each evaluation instance.
\textit{Our difficulty computation method is general and can be used with any other model or configurations; we use RoBERTa-large as it has been shown to achieve high performance across diverse NLP tasks} \cite{Liu2019RoBERTaAR}.
In addition, we show that difficulty scores computed using our procedure also generalize for other models (\ref{generalization_sec}).

% \Neeraj{Can mention about some analysis (histograms) of difficulty scores}.
% Note that we use a single model (RoBERTa-large) to compute the difficulty scores and show that they generalize for other models also (Section \ref{}).

We note that difficulty computation is not our primary contribution. Prior work \cite{swayamdipta-etal-2020-dataset, xu-etal-2020-curriculum} has explored different ways to achieve this. However, our approach uses $120$ predictions from models trained with different configurations for its computation and hence is more reliable.
Equipped with difficulty scores of evaluation instances, we now demonstrate five applications of ILDAE in the following sections.

\section{Efficient Evaluations}
\label{sec_efficient_evaluations}

\subsection{Problem Statement}
Success of BERT \cite{devlin-etal-2019-bert} has fostered development of several other pre-trained language models such as RoBERTa \cite{Liu2019RoBERTaAR}, XLNet \cite{NEURIPS2019_dc6a7e65}, DistilBERT \cite{Sanh2019DistilBERTAD}, ALBERT \cite{Lan2020ALBERT:}.
Though, it has resulted in the availability of numerous model options for a task, comparing the performance of such a large number of models has become computationally expensive and time-consuming.
For example, in real-world applications like online competitions, the naive approach that evaluates candidate models on the entire test dataset would be too expensive because they receive thousands of model submissions and contain a sizable number of evaluation instances.
Moreover, some applications also require additional evaluations to measure Out-of-Domain generalization and robustness making it even more expensive.
\textit{Can we make the evaluations \textbf{efficient}?}

\subsection{Solution}
We address the above question and explore if the performance of candidate models can be accurately compared with a carefully selected smaller subset of the evaluation dataset. 
Reducing the number of instances would save computational cost and make the evaluations efficient.
To this end, we propose an approach that selects evaluation instances based on their difficulty scores. We compare performance of candidate models only on these selected instances and show that without considerably impacting the result of evaluations, our approach saves computational cost and time.

\begin{figure}[t!]
    \centering
    \includegraphics[width=7cm]{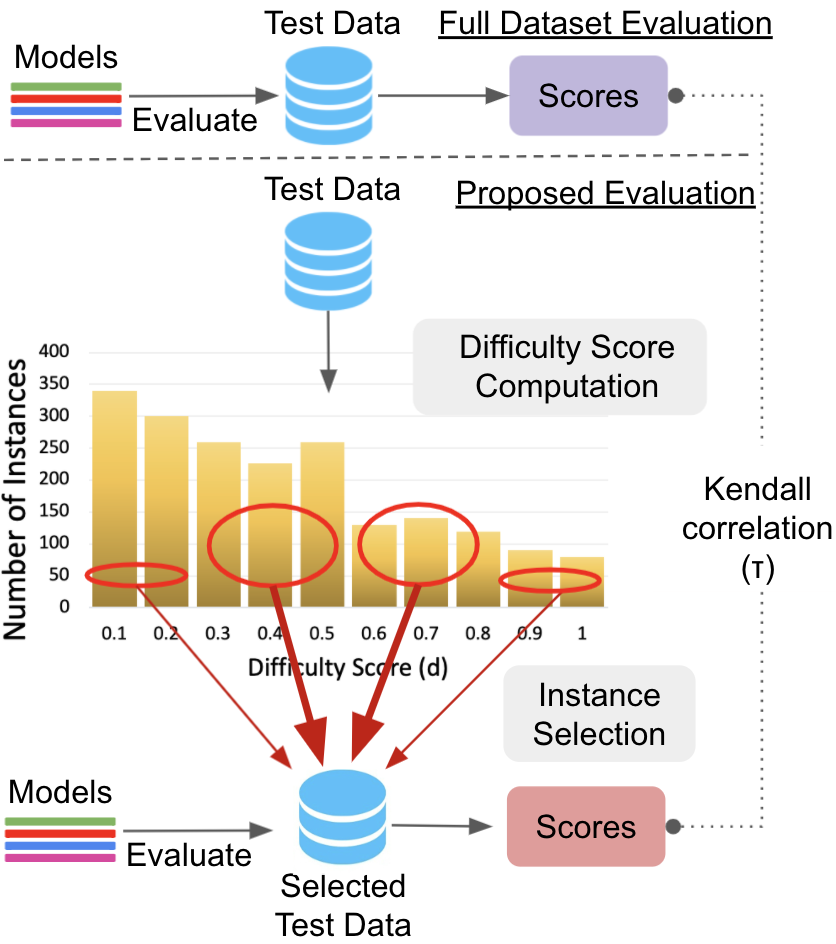}
    \caption{\textbf{Comparing standard evaluation approach (top) with our proposed `efficient' approach (bottom)}.
    We leverage difficulty scores to select a small subset of evaluation instances on which the performance of models can be efficiently compared.
    Our selected subset contains a majority of the instances with moderate difficulty scores and only a few with extreme difficulty scores.
    % that only uses a small subset of evaluation instances.
    % evaluates models only on the selected instances. 
    % Given an evaluation budget on the number of instances, we leverage the difficulty scores and
    % to select a small subset on which the candidate models get evaluated. 
    % Given a budget on the number of instances that can be evaluated, we 
    % select majority of the instances with moderate difficulty scores and only a few instances with very low or very high difficulty scores (depicted by area enclosed in red circles). 
    We use Kendall correlation between the performance scores to measure the efficacy of our approach.
    % Through our experiments, we show that these instances are sufficient to accurately compare the performance of models by measuring kendall correlation with the performance scores obtained on full dataset evaluation.
    }
    \label{fig:approach_figure}
\end{figure}

% while being computationally efficient
% without significantly compromising the accuracy of the comparison.
\textbf{Instance Selection:} We argue that \textit{the instances with extreme difficulty scores (very low and very high scores) would not be very effective in distinguishing between the candidate models}. 
This is because the former instances are trivial and would be answered correctly by many/all candidate models, while the latter ones are hard and would be answered correctly by only a few/none models.
Therefore, given a budget on the number of evaluation instances, we select a majority of them with moderate difficulty scores.
However, to distinguish amongst very weak and amongst very strong candidates, we also include a small number of instances with extreme difficulty scores.
Figure \ref{fig:approach_figure} illustrates our approach.

Note that our approach does not add any computational overhead during evaluations as the difficulty scores are pre-computed.
Furthermore, \textit{we do not compute separate difficulty scores for each candidate model as it would defy the sole purpose of `efficient' evaluations.}
Instead, we compute difficulty scores using only one model (RoBERTa-large) and exclude it from the list of candidate models for a fair evaluation of our approach.
For our instance selection approach to work in this setting, the difficulty scores should generalize for other models.
We empirically prove this generalization capability and demonstrate the efficacy of our efficient evaluations approach in \ref{efficient_results_sec}.
% Furthermore, 
% compare it with other heuristic based instance selection techniques.

\subsection{Experimental Details}
\begin{figure}[h]
    \centering
    \includegraphics[width=7cm]{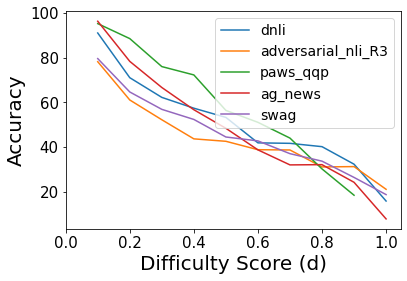}
    \caption{\textbf{Demonstrating difficulty score generalization.}
    Difficulty scores computed using RoBERTa-large show negative correlation with accuracy averaged over 27 other models, hence satisfying the desiderata mentioned in Section \ref{desiderate}.
    % The accuracy (averaged over 27 models) usually decreases with the increase in difficulty score (computed using RoBERTa-large only). 
    Note that we depict this trend for a few datasets only to avoid cluttering the image. Similar trend is observed for other dataset also\textsuperscript{\ref{footnote1}}.
    }
    \label{fig:difficulty_score_gen}
\end{figure}
% 

% \paragraph{Task:}
% The task of efficient evaluations boils down to selecting a subset of evaluation instances on which the performance of candidate models is compared.
% In contrast, the standard comparison approach uses the entire evaluation dataset (refer to Figure \ref{fig:approach_figure}).

\paragraph{Performance Metric:}
We measure the efficacy of an instance selection technique by computing accuracies of candidate models on the selected instances and calculating their Kendall's correlation \cite{kendall1938new} with accuracies obtained on the full evaluation dataset.
High correlation implies that the performance scores obtained using the selected instances display the same behavior as the performance scores obtained using the complete dataset. Hence, high correlations values are preferred.
% A high correlation is preferred as it corresponds to accurate evaluations.
% High correlation values are preferred as they indicate that evaluations can be accurately conducted even with fewer number of instances.

\paragraph{Datasets:} We experiment with a total of $23$ datasets across Natural Language Inference, Duplicate Detection, Sentiment Analysis, Question Answering, Commonsense Reasoning, and several other tasks.
Refer to Appendix section \ref{sec_datasets} for an exhaustive list of datasets for each task.
% such as SNLI \cite{bowman-etal-2015-large}, CommonsenseQA \cite{talmor-etal-2019-commonsenseqa}, QuaRel \cite{tafjord2019quarel}.

\paragraph{Candidate Models:}
\label{sec_models}
% We use RoBERTa-large \cite{Liu2019RoBERTaAR} model for computing the difficulty scores and 
We use BERT \cite{devlin-etal-2019-bert}, DistilBERT \cite{Sanh2019DistilBERTAD}, ConvBERT \cite{NEURIPS2020_96da2f59} , XLNET \cite{Yang2019XLNetGA}, SqueezeBERT \cite{iandola-etal-2020-squeezebert}, ELECTRA \cite{clark2020electra} in our experiments. 
We also use different variants of ConvBert (small, medium-small, base) and ELECTRA (small, base) models. 
For comprehensive experiments, we train each of the above models with training data of three different sizes ($2k$, $5k$, and $10k$ examples) resulting in $27$ candidate models for each dataset.
We intentionally exclude RoBERTa from this list as we use it for computing the difficulty scores.
\input{tables/efficient_eval_improved}

% \paragraph{Hyperparameters:}

% For our instance selection technique, we split the evaluation dataset into four regions after ordering them based on their difficulty scores and sample instances in the ratio $2: 3: 3: 2$ from these regions as shown by the area enclosed within red circles in Figure \ref{fig:approach_figure}. This ensures that majority of the instances with moderate difficulty score are sampled and 
% We also experiment with different hyperparameters and instance selection techniques, and show the results in supplementary section \ref{supp_sec_efficient_eval}.
% We use batch size of $32$ on Nvidia V100 16GB GPUs for our experiments. 

\paragraph{Instance Selection Baselines:}
We compare the proposed instance selection approach with the following baselines:

\textbf{Random Selection}:
Select a random subset of instances from the evaluation dataset.

\textbf{Heuristic Selection}:
Select instances based on the length heuristic (number of characters in the instance text) instead of the difficulty scores.
% We also experiment with different instance selection techniques using the difficulty scores and report the results in supplementary.

\subsection{Related Work}
Adaptive evaluation \cite{weiss1982improving} is used in educational settings for evaluating performance of students. It uses Item Response Theory (IRT) \cite{baker2004item} from psychometrics which requires a large number of subjects and items to estimate system parameters \cite{lalor-etal-2016-building,lalor-etal-2018-understanding}. Moreover, adaptive evaluation is computationally very expensive as it requires calculating the performance after each response and then selecting the next instance based on the previous responses of the subject.
Thus, it is not fit for our setting as we intend to improve the computational efficiency.
In contrast, our approach is much simpler and efficient as it does not incur any additional cost \underline{during} evaluation.
% \Neeraj{Provide evidence that it takes time and hence not practically feasible. It takes x additional time}

\subsection{Results}
\label{efficient_results_sec}
We first study generalization of our computed difficulty scores and then show the efficacy of the proposed instance selection approach in conducting efficient evaluations.

\subsubsection{Generalization of Difficulty Scores:}
\label{generalization_sec}
In Figure \ref{fig:difficulty_score_gen}, we plot accuracy (averaged over all 27 candidate models) against difficulty scores (computed using RoBERTa-large).
% If our difficulty scores (that are computed using a single model) generalize for other models then we expect them to be negatively correlated with accuracy (following the desiderata presented in Section \ref{desiderate}).
% In order to validate this, we plot accuracy (averaged across all 27 candidate models) against difficulty scores in Figure \ref{fig:difficulty_score_gen}. 
We find that with the increase in difficulty score, the accuracy consistently decreases for all datasets.
We also study this behavior for each individual candidate model 
% instead of averaging accuracy across all models 
and find results supporting the above observation\footnote{Further details are in appendix \label{footnote1}} (Figure \ref{fig:best_model}).
This proves that the difficulty scores follow the desiderata mentioned in Section \ref{desiderate} for other models also and our intuitions behind instance selection for conducting efficient evaluations hold true.
Note that these difficulty scores are computed using a specific model but our approach is general and will replicate this generalization capability if used with any other model.
% Note that our difficulty computation method is general and can be used with any model.
% \Neeraj{Should we say that our method should result in same observation but we leave that for future work}

% We use RoBERTa-large as it has been shown to achieve high performance across diverse NLP tasks.

\subsubsection{Efficient Evaluations:}
Table \ref{tab:efficiency_improved} shows Kendall correlation with full dataset evaluation achieved by various instance selection approaches for different percentages of instances.  

\paragraph{Proposed Approach Outperforms Baselines:}
Our proposed approach is consistently better than the Random and Heuristic approaches.
For instance, with just $0.5\%$ and $1\%$ evaluation instances, our approach outperforms the baseline methods by $\sim30\%$ and $\sim22.8\%$ respectively.
We show the expanded version of this table and performance of other instance selection techniques in Appendix.

\paragraph{Correlation Change with \% of Instances:}
As expected, Kendall correlation consistently increases as a higher percentage of instances are selected for evaluation. 
In case of SNLI, PAWS Wiki, QQP, DNLI, SWAG, and MNLI, just $2\%$ instances are sufficient to achieve correlation of $> 0.8$. 
For most datasets, with just $20\%$ of the evaluation instances, our approach achieves Kendall correlation of $>0.8$.
This suggests that the evaluations can be conducted with fewer instances without significantly compromising the accuracy of comparison. 
We further analyze performance of our approach for higher percentage of instances in Table \ref{tab:supp_efficiency_improved}.

Thus, for practical settings where candidate models can't be compared on the entire dataset due to computational and time constraints, evaluating only on the selected instances can result in fairly accurate performance comparison.

\paragraph{Performance on Multiple-Choice QA datasets:} 
Though, we perform better than the baselines approaches on almost all datasets, we achieve a lower correlation value for multiple-choice question answering datasets such as QuaRel, QuaRTz, and Winogrande.
% than most classification datasets (such as SNLI, MNLI, DNLI, QNLI, QQP).
We attribute this behavior to the close scores (accuracies) achieved by many candidate models even in case of full dataset evaluation. 
Thus, it is difficult to differentiate such models as they achieve nearly the same performance.
% This is observed for both full dataset evaluation and evaluation on selected instances.
Furthermore, in some difficult datasets such as Adversarial NLI (R1, R2, and R3), ARC Difficult, and Winogrande, many candidate models achieve accuracies very close to the random baseline ($33\%$ for NLI, $50\%$ for Winogrande). 
So, comparing their performance even with full dataset does not provide any significant insights.

\section{Improving Evaluation Datasets}
\label{sec_evolving_instances}

\subsection{Problem Statement}
Recent years have seen a rapid increase in the number and size of NLP datasets. 
Crowd-sourcing is a prominent way of collecting these datasets.
Prior work \cite{gururangan-etal-2018-annotation,Tan2019InvestigatingBI, mishra2020dqi} has shown that crowd-sourced datasets can contain:
(a) \textit{erroneous instances} that have annotation mistakes or ambiguity,  
(b) too many \textit{trivial instances} that are very easy to answer.
% since creating such instances takes much less effort on part of the crowd-worker.
This hampers the quality of the dataset and makes it less reliable for drawing conclusions.
\textit{Can difficulty scores aid in \text{improving} the quality of evaluation datasets?}

% We define evolution to be something that rectifies the previous mistakes, eliminates the useless instances and present more challenging instances.
% As stronger models come up, the evaluation datasets saturate.
% they are bound to contain a few erroneous instances (mislabeled or with ambiguous answer) 
% dynabench or adversarial NLI presents one way of creating challenging instances. 
% But it results in increasing the size of evaluation dataset and also requires creating the new example from scratch.
\input{tables/erroneous_examples}
\subsection{Solution}
We first show that erroneous and trivial instances can be identified using the difficulty scores and then present a human-and-model-in-the-loop technique to modify/repair such instances resulting in improved quality of the datasets.

% In this subsection, we first analyze the difficulty scores and show that they can be used to identify erroneous and trivial instances.

\paragraph{Identifying Erroneous and Trivial Instances: }
% We hypothesize that the instances that get assigned very high difficulty score might contain some erroneous instances because even the strongest of the models fail to answer them correctly.
We inspect $50$ instances each with very high and very low difficulty scores and find that a significant percentage of the former are either mislabeled or contain ambiguity and the latter are too easy to be answered.
% This is intuitive because the instances with very high difficulty score are answered incorrectly by many/all models with very high confidence. 
% Thus, 

% \Neeraj{This is intuitive because...}
% \Neeraj{Say that this has also been explored in prior work but we do something extra}
% This is intuitive because the instances with very high difficulty scores could be erroneous because nearly all the models fail to answer them correctly and the instances with very low difficulty scores could be trivial because nearly all the models answer them correctly with high confidence.

Table \ref{tab:erroneous_examples} shows examples of erroneous instances from SNLI, Winogrande, CSQA, and Abductive NLI.
We find $72\%$ of the inspected SNLI instances to be erroneous.
Furthermore, we find that some high difficulty score instances are actually difficult even for humans because they require abilities such as commonsense reasoning.
Table \ref{tab:supp_actually_difficult_examples} (appendix) shows such instances.
We also provide examples of trivial instances (Table \ref{tab:trivial_examples}) and note that such instances are trivial from model's perspective as they can be answered correctly (with high confidence) by simply latching on to some statistical cues present in the training data.

\begin{figure}
    \centering
    \includegraphics[width=5.3cm]{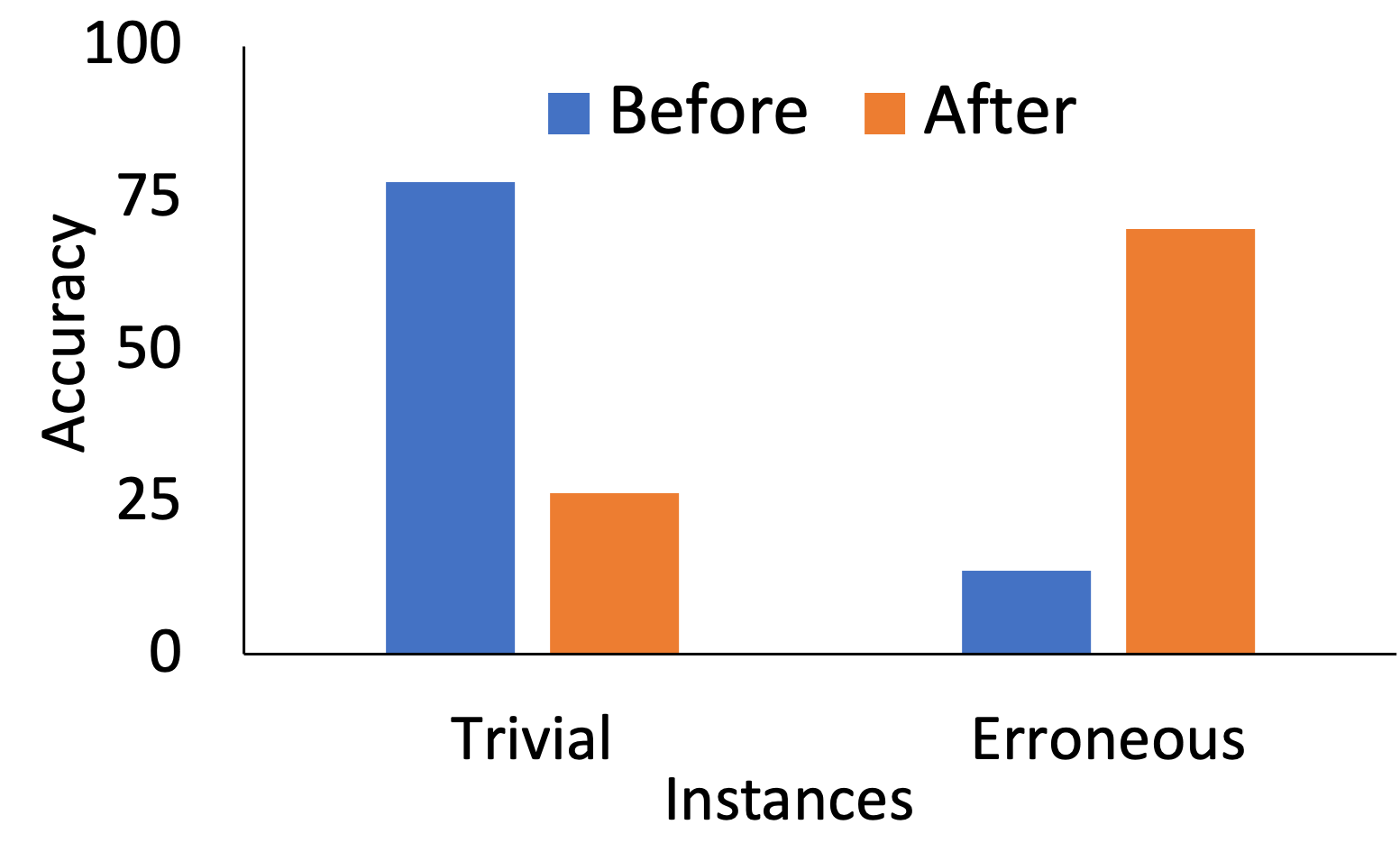}
    \caption{Comparing \textbf{accuracy} (averaged over $27$ models) \textbf{before and after modifying the SNLI instances} using our model-and-human-in-the-loop technique.
    % averaged accuracy of $27$ models on trivial and erroneous instances before and after modifying them using our model-and-human-in-the-loop technique. 
    The accuracy on trivial instances decreases as we make them more difficult while the accuracy on erroneous instances increases as we repair them.
    }
    \label{fig:modified_accuracy}
\end{figure}
\paragraph{Technique: }
Since the trivial instances are too easy to be answered, we propose to modify them in an \textit{adversarial} way such that they no longer remain trivial. 
Specifically, we include a human-in-the-loop who needs to modify a trivial instance in a label-preserving manner such that the modified version fools the model into making an incorrect prediction.
For adversarial attack, we use the strongest model from our ensemble of $120$ models. 
It has two key differences with the standard adversarial data creation approach presented in \cite{nie-etal-2020-adversarial, kiela-etal-2021-dynabench}:
(a) it requires modifying an already existing instance instead of creating a new instance from scratch.
(b) it does not increase the size of the evaluation dataset as we replace an already saturated instance (trivial) with its improved not-trivial version. 
We use a human instead of leveraging automated ways to modify the trivial instances because our objective is to improve the quality of instances and prior work has shown that these automated techniques often result in unnatural and noisy instances.
Therefore, such techniques could be cost-efficient but might not solve the sole purpose of improving quality.

To further improve the quality, we provide instances with very high difficulty score (potentially erroneous) and ask a human to repair them such that the repaired versions follow the task definition.
The human can either change the instance text or its answer to achieve the goal.
Note that this scenario is model-independent. 

% For the second scenario, it provides potentially erroneous instances (those with very high difficulty score) with their originally annotated 
% and the most probable 
% answer, and the human needs to repair either the instance text (if it is ambiguous) or the answer (if it is mislabeled) such that the repaired version follows the task definition. 
% This scenario does not involve the model as it simply requires repairing the instances.
% Note that the first scenario is an iterative process as it involves fooling the model and the second does not involve the model.
% Figure \ref{} illustrates both these scenarios.
% \Neeraj{Add Figure}

\subsection{Results}

% We instantiate this framework for SNLI dataset.
Table \ref{tab:modified_examples} shows original and modified instances from SNLI.
Top two examples correspond to the trivial instances where the human modified the hypothesis in a label-preserving manner such that it fooled the model into making incorrect prediction.
The bottom two correspond to the mislabeled instances where the human rectified the label.
Figure \ref{fig:modified_accuracy} compares the performance of models on the original instances and the their modified/repaired versions. 
As expected, the performance drops on the previously trivial instances as they are no longer trivial and improves on the previously erroneous instances.
We release the improved version of the dataset compiled via our technique.

\input{tables/modified_examples}
\section{Other Applications of ILDAE}
\label{sec_other_applications}
We now briefly discuss other ILDAE applications.

% In this section, we discuss other applications of ILDAE.
\begin{figure}
    \centering
    \includegraphics[width=6.4cm]{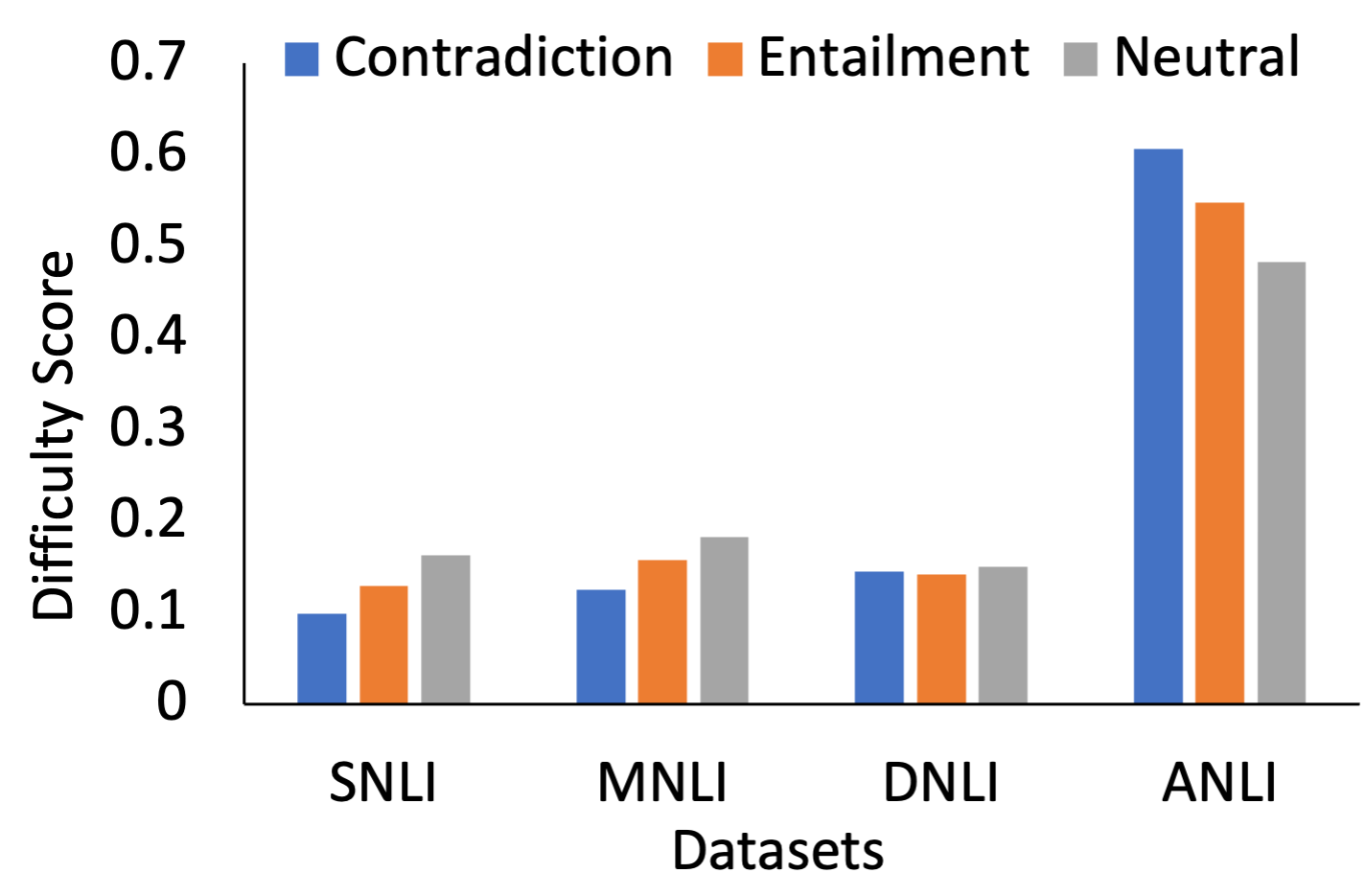}
    \caption{Comparing \textbf{average difficulty of NLI datasets} for each label (Entailment, Contradiction, and Neutral). 
    % This reveals characteristics of datasets such as for SNLI and MNLI, models find it easier to identify contraction instances as compared to entailment and neutral while for Adversarial NLI, it is the opposite.
    }
    \label{fig:nli_label_analysis}
\end{figure}
\subsection{Dataset Analysis}
\label{sec_dataset_analysis}
ILDAE reveals several useful characteristics of datasets such as which class label has the easiest instances.
We study this for NLI datasets: SNLI, MNLI, DNLI, and Adversarial NLI (Figure \ref{fig:nli_label_analysis}). 
For SNLI and MNLI, we find that the contradiction instances receive lower average difficulty score than entailment and neutral instances.
For Adversarial NLI, the order is reversed. 
For DNLI, all the labels get assigned nearly the same average difficulty.
% This indicates that for SNLI and MNLI datasets, models find it easier to identify contradiction instances as compared to entailment and neutral instances but for ANLI it is the opposite.
% This indicates that models find it easier to identify a contradiction example as compared to entailment and neutral.
% Note that this is a characteristic of these datasets and not necessarily a characteristic of the task.
% This implies that the trend could be different for some other NLI dataset.
Such analysis can serve as a guide for future data creation as it indicates for which type of instances more data collection effort needs to be invested. 
It can also be used to compare average difficulty at dataset level. 
Furthermore, a new harder task-specific benchmark can be created by combining high difficulty instances from all the datasets of that task.

% \begin{figure*}[t]
% \centering
%     \begin{subfigure}{.4\linewidth}
%         \includegraphics[width=\linewidth]{Pictures/best_model_snli.png}
%         \caption{SNLI}
%     \end{subfigure}
%     \begin{subfigure}{.4\linewidth}
%          \includegraphics[width=\linewidth]{Pictures/best_model_dnli.png}
%          \caption{DNLI}
%     \end{subfigure}
%     % \begin{subfigure}{.32\linewidth}
%     %      \includegraphics[width=\linewidth,height=27mm]{Pictures/best_model.png}
%     %      \caption{Dataset C}
%     % \end{subfigure}
    
%     \caption{Variation of accuracy of each candidate model with difficulty score for SNLI and DNLI datasets. Each line corresponds to a candidate model (27 in total). It shows that a single model does not achieve the highest accuracy across all difficulty scores. Thus, model selection depends on the target scenario. For instance, if the target task is difficult then the model that achieves highest performance on instances with high difficulty score should be selected.}
%     \label{fig:best_model}    
% \end{figure*}

\subsection{Model Analysis}
\label{sec_model_analysis}

\begin{figure}
    \centering
    \includegraphics[width=7cm,height=6cm]{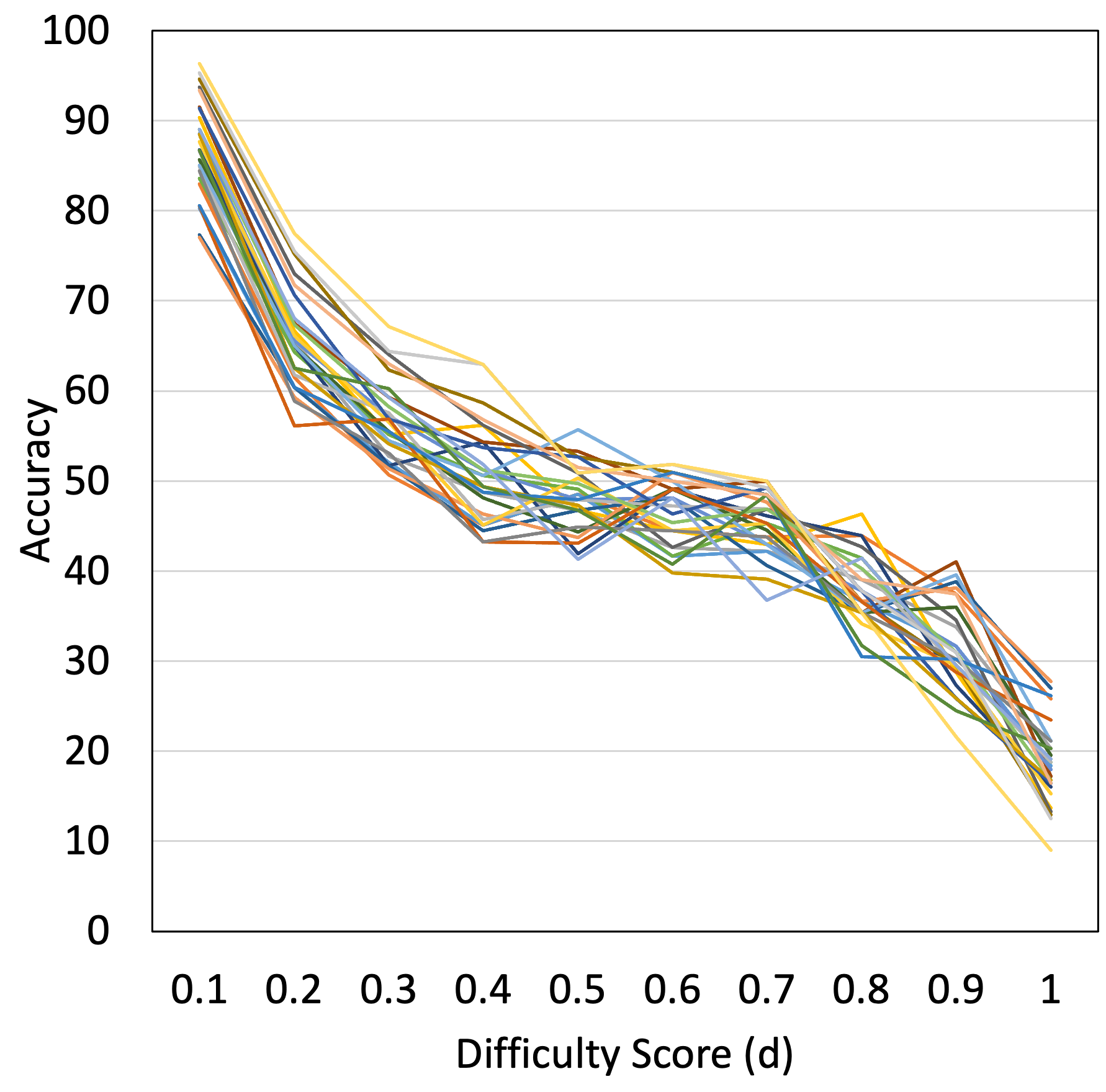}
    \caption{Variation of accuracy of each candidate model with difficulty score for SNLI dataset. Each line corresponds to a candidate model (27 in total). \textbf{It shows that a single model does not achieve the highest accuracy across all difficulty scores}. 
    % Thus, model selection depends on the target scenario. For instance, if the target task is difficult then the model that achieves highest performance on instances with high difficulty score should be selected.
    }
    \label{fig:best_model}
\end{figure}

We divide the evaluation instances into different regions based on the difficulty scores and analyze models' performance in each region.
We find that a single model does not achieve the highest accuracy across all regions.
Figure \ref{fig:best_model} illustrates this pattern for SNLI dataset.
This implies that the model that achieves the highest performance on easy instances may not necessarily achieve the highest performance on difficult instances.
The similar pattern is observed for other datasets (refer appendix).
Such analysis would benefit in model selection. 
For instance, in scenarios where a system is expected to encounter hard instances, we can select the model that has the highest accuracy on instances of difficult regions. Whereas, for scenarios containing easy instances, the model that has the highest accuracy on instances of easy regions.

% \textit{Does a single model achieve the highest performance across all difficulty scores?}

% Thus, model selection depends on the application scenario.
% For instance, if generalization is preferred then the model that achieves higher performance on difficult instances should be selected,
% if the task is expected to contain easy instances only then the model that achieves higher performance on the instances with low difficulty score should be selected.
% Hence, difficulty scores can help in selecting the best-fit model for specific scenarios.

% \subsection{Weighted Accuracy Shows Better Correlation with OOD Performance}
\subsection{Correlation with OOD Performance}
\label{sec_weighted_acc}
Large pre-trained language models can achieve high In-Domain performance on numerous tasks. However, it does not correlate well with OOD performance \cite{hendrycks2018benchmarking, hendrycks-etal-2020-pretrained}.
To this end, we present an approach to compute a weighted accuracy that shifts away from treating all the evaluations instances equally and assigns weight based on their difficulty scores.
We define the weight $w_i$ of an instance $i$ with difficulty score $d_i$ as:
\[
     w_i = \frac{1 + \mu*d_i}{N + \mu*\sum_{j=1}^{N}d_j}
\]
%  \sum_{j=1}^{E} c_{ji}
where $N$ corresponds to the total number of evaluation instances, and $\mu$ is a hyper-parameter that controls influence of difficulty score on the weight.
Then, weighted accuracy $W$ is simply:

\[
     W = \sum_{i=1}^{N}w_i*v_i
\]

where $v_i$ is 1 when the model's prediction is correct else 0.
This implies that high accuracy may not always translate to high weighted accuracy.

\begin{figure}[t!]
    \centering
    \includegraphics[width=6cm]{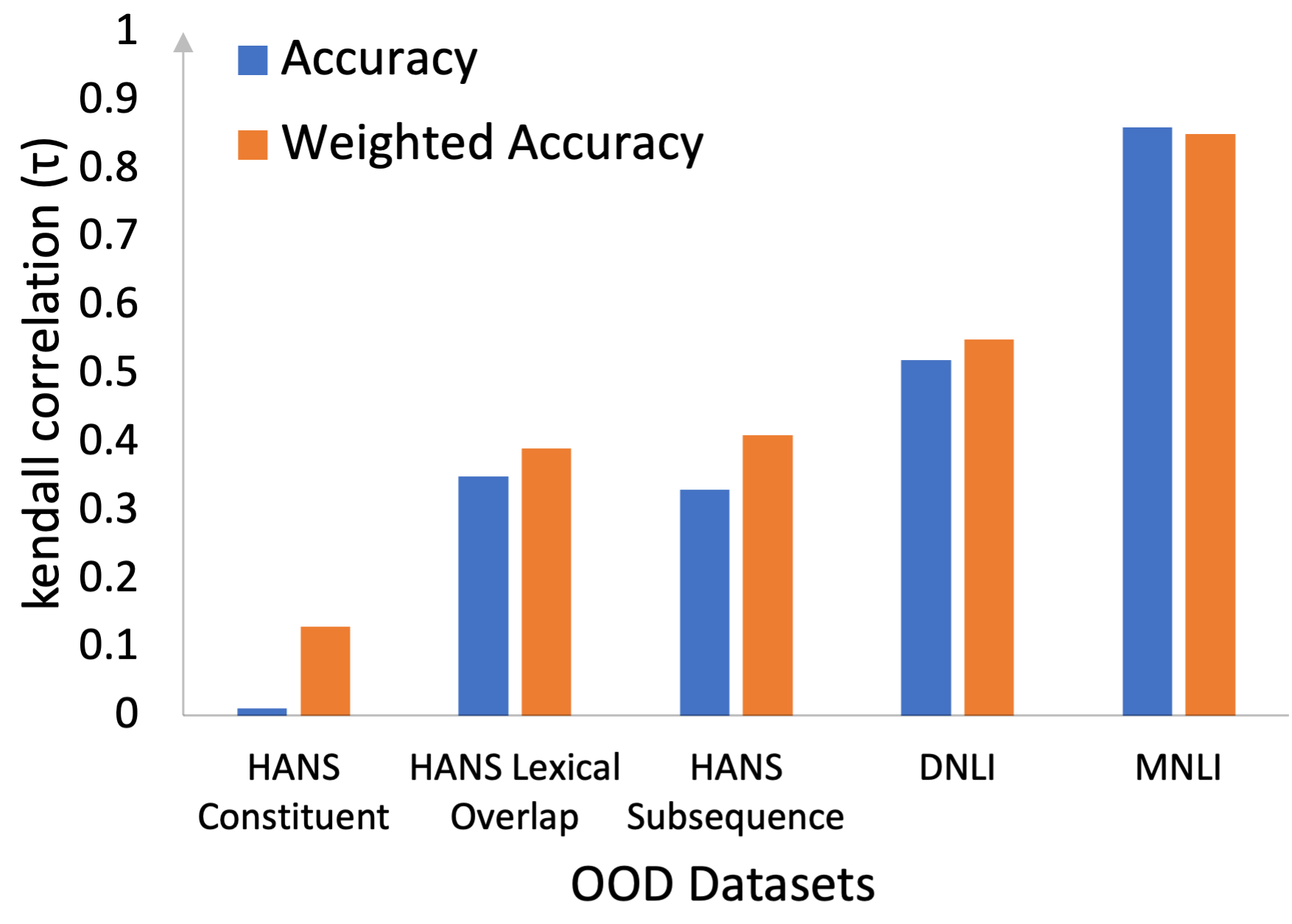}
    \caption{Comparing Kendall correlation of standard unweighted accuracy and weighted accuracy with OOD accuracy. 
    % We use SNLI as the in-domain dataset and HANS (Constituent, Lexical Overlap, Subsequence), DNLI, MNLI as OOD datasets. 
    \textbf{Weighted accuracy achieves $5.2\%$ higher correlation on average.}}
    \label{fig:ood_correlation}
\end{figure}

We take SNLI as the in-domain dataset and MNLI, DNLI, and HANS \cite{mccoy-etal-2019-right} (Constituent, Lexical Overlap, Subsequence) as OOD datasets.
We calculate unweighted and weighted accuracy of the 27 models (described in Section \ref{sec_models}) and compare their Kendall correlation with the accuracy on OOD datasets.
Figure \ref{fig:ood_correlation} shows this comparison.
It can be observed that weighted accuracy shows $5.2\%$ higher correlation with OOD performance.
Most improvement is observed in datasets that have many instances with high difficulty score (HANS).
Thus, weighting instances based on their difficulty score is more informative than the standard accuracy.

\section{Conclusion}
We conducted Instance-Level Difficulty Analysis of Evaluation data (ILDAE) in a large-scale setup of $23$ datasets and presented its five novel applications.
With these applications, we demonstrated ILDAE's impact in several important areas, such as conducting efficient evaluations with fewer instances, improving dataset quality, and estimating out-of-domain performance reliably. 
% Several empirical results associated with ILDAE such as (1) reducing the evaluation instances to 0.2x and still outperforming the evaluation with full dataset by a Kendall correlation of $0.72$, (2) Improving OOD performance evaluation by $5.2\%$ Kendall correlation just by using weighted evaluation metric (without using any OOD dataset) have shown promise for a landscape of several new applications associated with ILDAE. 
We release our computed difficulty scores and hope that our analyses and findings will bring more attention to this important yet underexplored field of leveraging instance difficulty in evaluations.

\section*{Ethical Considerations}  
We use existing public-domain text datasets, such as SNLI, Winogrande, and ARC, and follow the protocol to use and adapt research data to compute instance-level difficulty scores. We will release the computed difficulty scores, but will not share the original source data.  We recommend readers to refer to the original source research papers. Any bias observed in difficulty scores computed using our methods can be attributed to the source data and our computation functions. However, no particular socio-political bias is emphasized or reduced specifically by our methods. 

\bibliography{anthology,custom}
\bibliographystyle{acl_natbib}
\newpage
\appendix

\section{Difficulty Score Analysis}
\label{sec_diff_score_gen_supp}

\subsection{Generalization}

\begin{figure}[h]
    \centering
    \includegraphics[width=8cm]{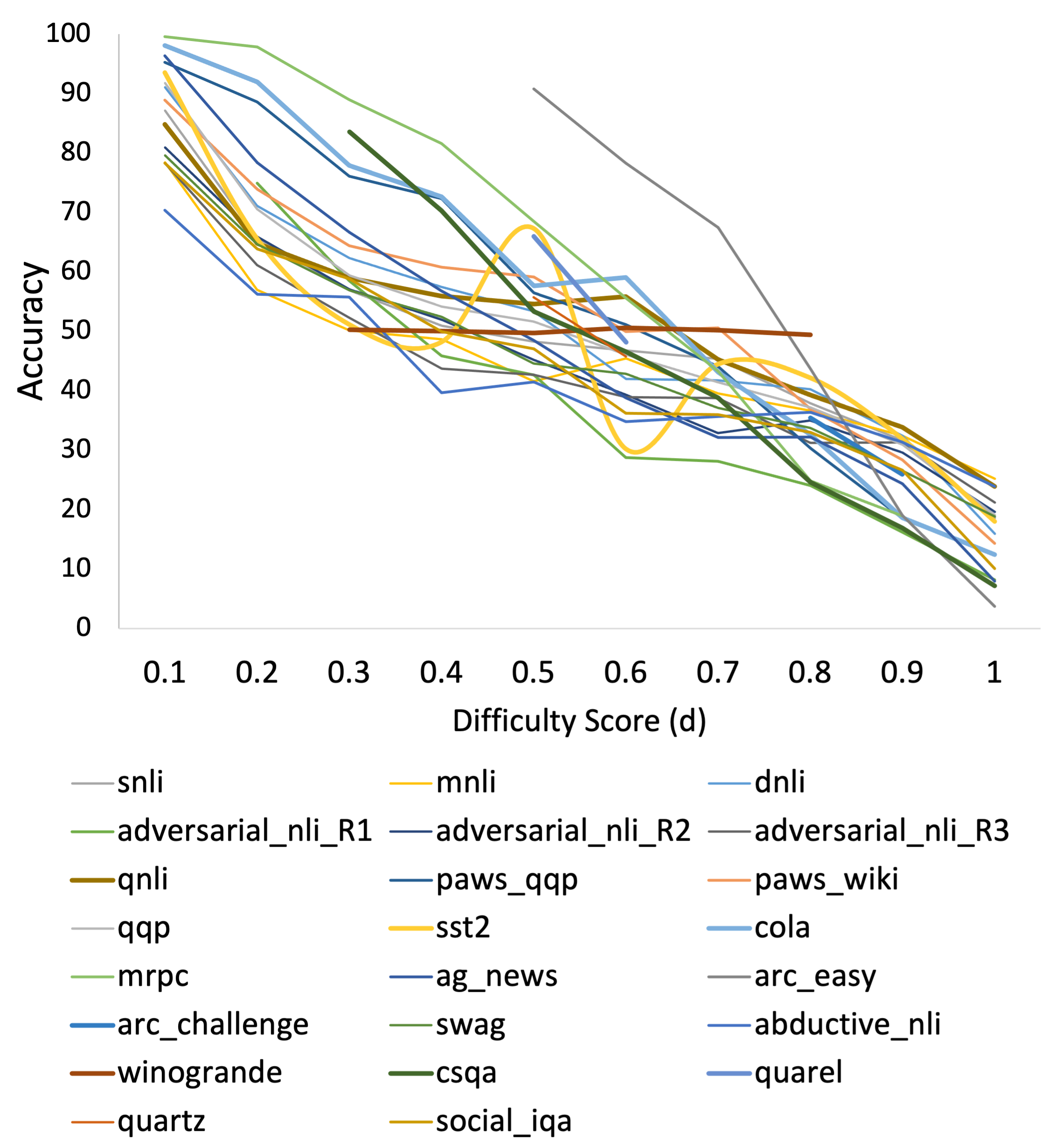}
    \caption{Demonstrating difficulty score generalization. It shows the variation of accuracy (averaged over 27 models) with difficulty scores (computed using RoBERTa-large only). The accuracy usually decreases with the increase in difficulty proving the generalization capability of our difficulty scores.}
    \label{fig:supp_difficulty_score_gen}
\end{figure}

Figure \ref{fig:supp_difficulty_score_gen} shows the trend of accuracy with difficulty scores.

\subsection{Difficulty Score Vs Accuracy}

\begin{figure*}[t]
\centering
    \begin{subfigure}{.4\linewidth}
        \includegraphics[width=\linewidth]{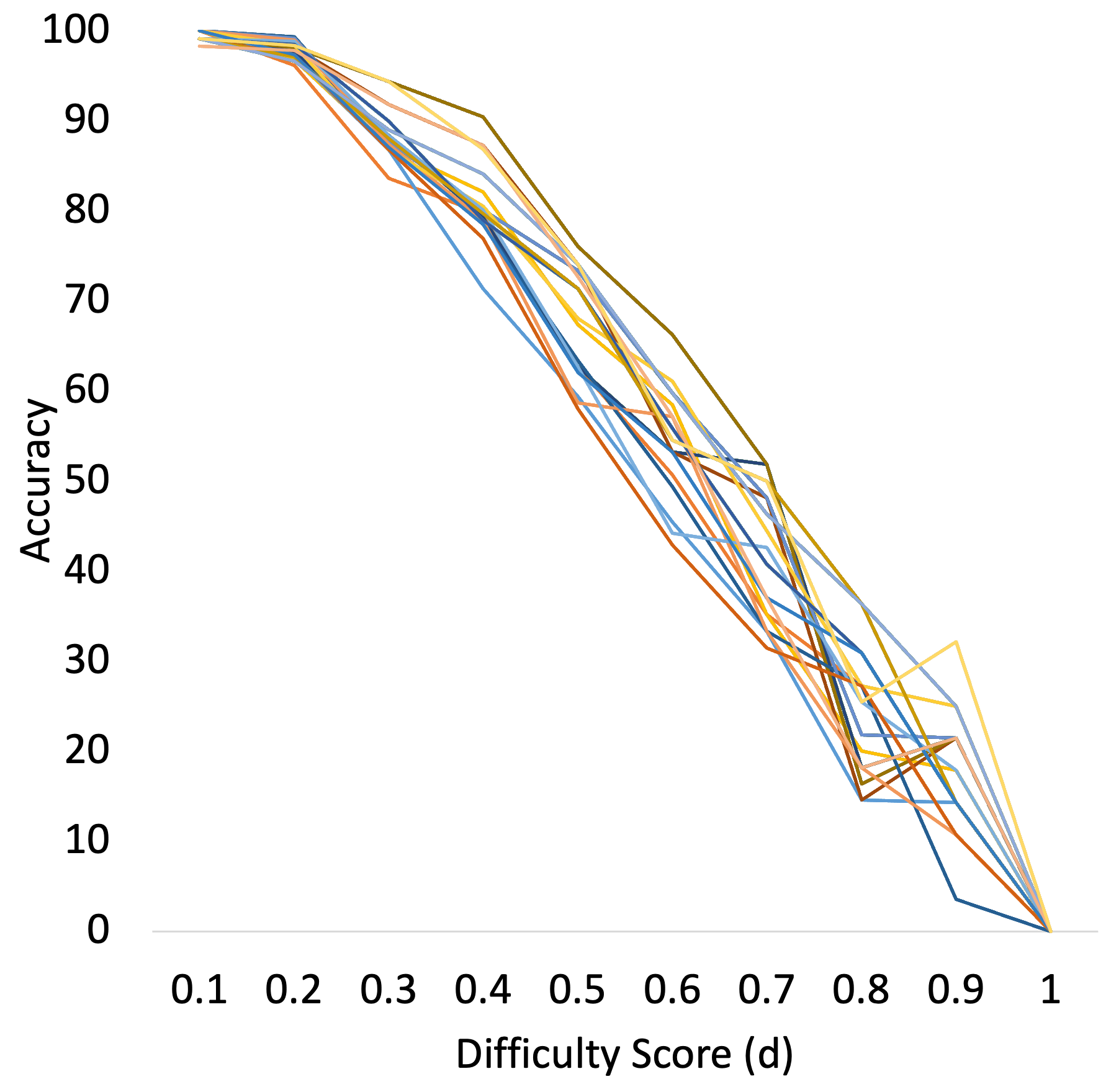}
        \caption{MRPC}
    \end{subfigure}
    \begin{subfigure}{.4\linewidth}
         \includegraphics[width=\linewidth]{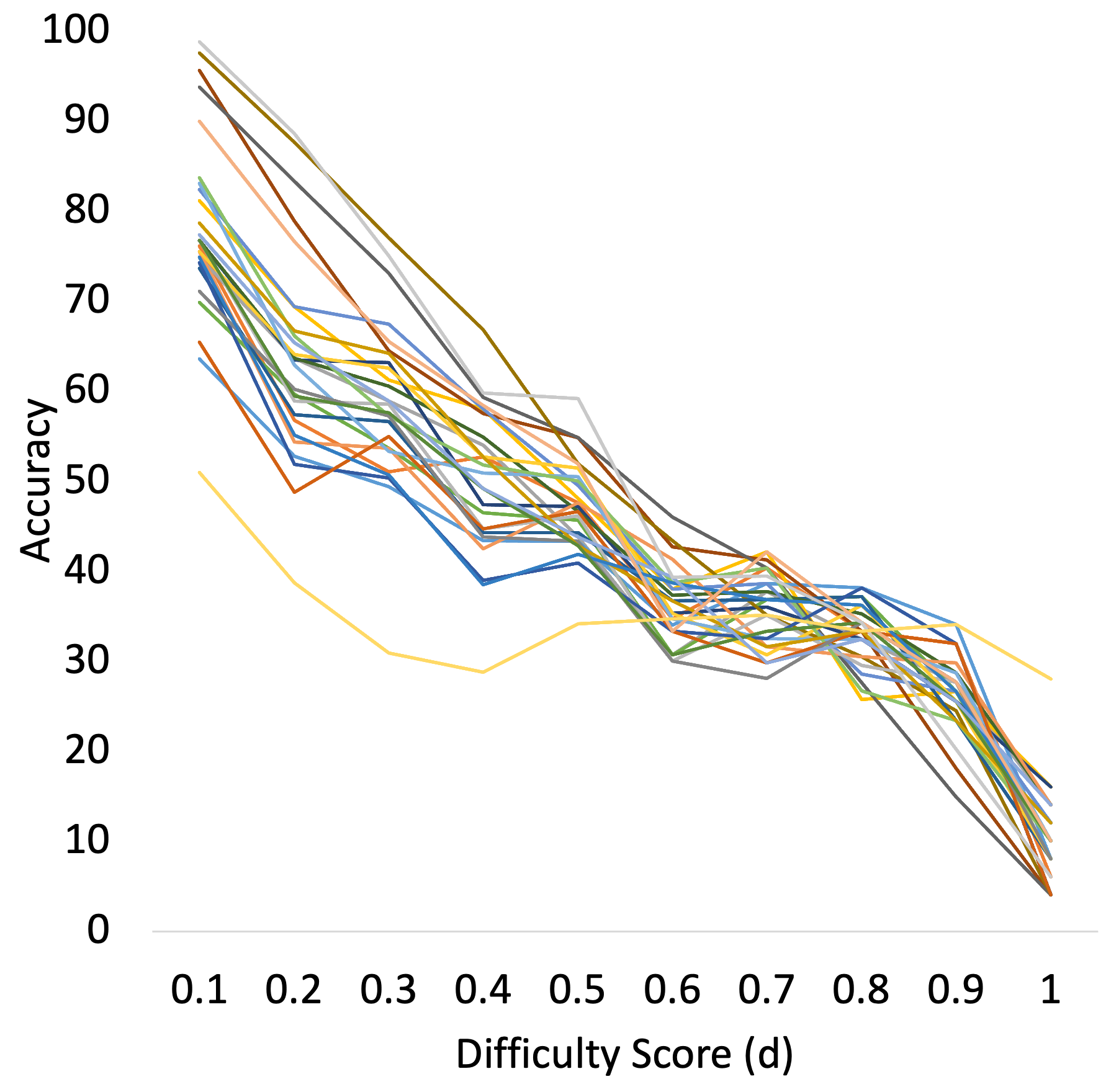}
         \caption{SocialIQA}
    \end{subfigure}
    % \begin{subfigure}{.32\linewidth}
    %      \includegraphics[width=\linewidth,height=27mm]{Pictures/best_model.png}
    %      \caption{Dataset C}
    % \end{subfigure}
    
    \caption{Variation of accuracy with difficulty score for MRPC and SocialIQA datasets. Each line corresponds to a candidate model (27 in total).}
    \label{fig:supp_best_model}    
\end{figure*}

Figure \ref{fig:supp_best_model} shows the trend of accuracy against difficulty scores for each individual model for MRPC and SocialIQA datasets. 

\section{Datasets:}
\label{sec_datasets}
We experiment with the following datasets:
SNLI \cite{bowman-etal-2015-large}, Multi-NLI \cite{williams-etal-2018-broad}, Dialogue NLI \cite{welleck-etal-2019-dialogue}, Adversarial NLI (R1, R2, R3) \cite{nie-etal-2020-adversarial}, QNLI \cite{wang-etal-2018-glue}, QQP \cite{iyer2017first}, MRPC \cite{dolan2005automatically}, PAWS-QQP, PAWS-Wiki \cite{zhang-etal-2019-paws}, SST-2 \cite{socher-etal-2013-recursive}, COLA \cite{warstadt-etal-2019-neural} AG's News \cite{NIPS2015_250cf8b5}, ARC-Easy, ARC-Challenge \cite{clark2018think}, SWAG \cite{zellers2018swagaf}, Abductive-NLI \cite{bhagavatula2020abductive}, Winogrande \cite{Sakaguchi2020WINOGRANDEAA}, CommonsenseQA \cite{talmor-etal-2019-commonsenseqa}, QuaRel \cite{tafjord2019quarel}, QuaRTz \cite{tafjord-etal-2019-quartz}, and SocialIQA \cite{Sap2019SocialIC}.

\section{Efficient Evaluations}
\label{supp_sec_efficient_eval}
Table \ref{tab:supp_efficiency_improved} shows the Kendall correlation with full dataset evaluation achieved by our  instance selection approach for different percentages of instances.  
\input{tables/efficient_eval_supp}

\section{Actually Difficult Instances}
\input{tables/actually_difficult_examples}
Table \ref{tab:supp_actually_difficult_examples} shows examples of instances that get assigned very high difficulty score but are actually difficult even for humans because they require reasoning abilities such as commonsense knowledge.

\section{Erroneous Instances}

\input{tables/supp_erroneous}
Table \ref{tab:supp_erroneous_examples} shows examples of erroneous instances.

\section{Trivial Instances}
\input{tables/trivial_examples}
Table \ref{tab:trivial_examples} shows examples of trivial instances.
% \appendix

% \section{Example Appendix}
% \label{sec:appendix}

% This is an appendix.

\end{document}

%% file: tables/efficient_eval_improved.tex
\begin{table*}[]
\centering
\resizebox{\textwidth}{!}{%
\begin{tabular}{l|lll|lll|l|l|l|l}
\toprule
\textbf{\% Instances $\rightarrow$} &
\multicolumn{3}{c}{\textbf{0.5\%}} &
\multicolumn{3}{c}{\textbf{1\%}} &
\multicolumn{1}{c}{\textbf{2\%}} &
\multicolumn{1}{c}{\textbf{5\%}} &
\multicolumn{1}{c}{\textbf{10\%}} &
\multicolumn{1}{c}{\textbf{20\%}} \\
\textbf{Dataset $\downarrow$} & \textbf{Random} & \textbf{Heuristic} & \textbf{Proposed} & \textbf{Random} & \textbf{Heuristic} & \textbf{Proposed} & \textbf{Proposed} & \textbf{Proposed} & \textbf{Proposed} & \textbf{Proposed}  \\

\midrule
% order - paws qqp, qqp, dnli, paws wiki cola, swag, snli, mnli, adv nli, qnli, sst2, rte, arpc, agnews, emotions, arc, abductive, winogrande, csqa, quarel, quartz, socialiqa 

% snli, paws wiki, adv nli R1, ag news, qnli, mrpc, social iqa, 

SNLI & $0.55_{0.09}$ & $0.38_{0.17}$ & $\textbf{0.68}_{0.13}$ & $0.68_{0.05}$ & $0.58_{0.08}$ & $\textbf{0.78}_{0.08}$ & $0.83_{0.04}$ & $0.88_{0.04}$ & $0.91_{0.01}$ & $0.93_{0.02}$ \\ 

PAWS Wiki & $0.67_{0.07}$ & $0.68_{0.04}$ & $\textbf{0.78}_{0.06}$ & $0.73_{0.05}$ & $0.78_{0.02}$ & $\textbf{0.86}_{0.05}$ & $0.89_{0.02}$ & $0.91_{0.03}$ & $0.95_{0.01}$ & $0.96_{0.01}$ \\ 

AgNews & $0.12_{0.26}$ & $0.14_{0.27}$ & $\textbf{0.47}_{0.05}$ & $0.25_{0.34}$ & $0.41_{0.14}$ & $\textbf{0.52}_{0.1}$ & $0.65_{0.07}$ & $0.75_{0.06}$ & $0.8_{0.04}$ & $0.89_{0.03}$ \\ 

QNLI & $0.41_{0.1}$ & $0.44_{0.04}$ & $\textbf{0.48}_{0.13}$ & $0.57_{0.04}$ & $0.55_{0.1}$ & $\textbf{0.57}_{0.07}$ & $0.7_{0.06}$ & $0.78_{0.06}$ & $0.85_{0.03}$ & $0.91_{0.03}$ \\ 

MRPC & $0.04_{0.09}$ & $-0.03_{0.18}$ & $\textbf{0.21}_{0.16}$ & $-0.02_{0.09}$ & $0.05_{0.2}$ & $\textbf{0.29}_{0.21}$ & $0.36_{0.15}$ & $0.45_{0.08}$ & $0.58_{0.12}$ & $0.65_{0.14}$ \\ 

SocialIQA & $0.19_{0.09}$ & $0.15_{0.29}$ & $\textbf{0.37}_{0.17}$ & $0.34_{0.07}$ & $0.28_{0.21}$ & $\textbf{0.4}_{0.09}$ & $0.58_{0.1}$ & $0.67_{0.04}$ & $0.75_{0.08}$ & $0.81_{0.05}$ \\

QQP & $0.63_{0.06}$ & $0.64_{0.05}$ & $\textbf{0.65}_{0.05}$ & $0.74_{0.03}$ & $0.74_{0.01}$ & $\textbf{0.77}_{0.06}$ & $0.84_{0.04}$ & $0.9_{0.04}$ & $0.94_{0.04}$ & $0.95_{0.01}$ \\ 

DNLI & $0.58_{0.05}$ & $0.59_{0.1}$ & $\textbf{0.58}_{0.11}$ & $0.68_{0.1}$ & $0.71_{0.04}$ & $\textbf{0.76}_{0.07}$ & $0.84_{0.04}$ & $0.92_{0.05}$ & $0.94_{0.03}$ & $0.96_{0.01}$ \\

COLA & $-$ & $-$ & $-$ & $-0.01_{0.18}$ & $\textbf{0.25}_{0.26}$ & $0.24_{0.45}$ & $0.41_{0.41}$ & $0.63_{0.23}$ & $0.75_{0.08}$ & $0.78_{0.02}$ \\ 

SWAG & $0.72_{0.04}$ & $0.66_{0.02}$ & $\textbf{0.75}_{0.06}$ & $0.79_{0.03}$ & $0.77_{0.03}$ & $\textbf{0.78}_{0.05}$ & $0.86_{0.03}$ & $0.89_{0.02}$ & $0.93_{0.01}$ & $0.95_{0.01}$ \\ 

PAWS QQP & $-$ & $-$ & $-$ & $0.13_{0.24}$ & $0.36_{0.05}$ & $\textbf{0.34}_{0.13}$ & $0.55_{0.19}$ & $0.8_{0.05}$ & $0.84_{0.03}$ & $0.87_{0.04}$ \\

MNLI & $0.7_{0.04}$ & $0.71_{0.03}$ & $\textbf{0.73}_{0.07}$ & $0.8_{0.02}$ & $0.8_{0.04}$ & $\textbf{0.82}_{0.08}$ & $0.89_{0.03}$ & $0.93_{0.02}$ & $0.95_{0.02}$ & $0.96_{0.01}$ \\ 

Adv. NLI R1 & $0.0_{0.08}$ & $-0.07_{0.06}$ & $\textbf{0.17}_{0.27}$ & $0.02_{0.13}$ & $\textbf{0.09}_{0.11}$ & $0.08_{0.2}$ & $0.13_{0.18}$ & $0.3_{0.18}$ & $0.47_{0.05}$ & $0.59_{0.05}$ \\ 

Adv. NLI R2 & $-0.08_{0.04}$ & $\textbf{-0.01}_{0.06}$ & $-0.08_{0.16}$ & $-0.08_{0.07}$ & $\textbf{0.02}_{0.03}$ & $-0.03_{0.21}$ & $0.0_{0.12}$ & $0.17_{0.03}$ & $0.26_{0.11}$ & $0.42_{0.15}$ \\ 

Adv. NLI R3 & $-0.15_{0.12}$ & $\textbf{0.15}_{0.1}$ & $0.1_{0.21}$ & $-0.03_{0.06}$ & $0.07_{0.1}$ & $\textbf{0.1}_{0.11}$ & $0.18_{0.16}$ & $0.12_{0.17}$ & $0.31_{0.15}$ & $0.58_{0.05}$ \\ 

SST-2 & $-$ & $-$ & $-$ & $0.08_{0.15}$ & $0.16_{0.35}$ & $\textbf{0.29}_{0.25}$ & $0.4_{0.2}$ & $0.52_{0.16}$ & $0.65_{0.13}$ & $0.81_{0.08}$ \\

ARC Easy & $-$ & $-$ & $-$ & $0.0_{0.2}$ & $-0.03_{0.12}$ & $\textbf{0.42}_{0.19}$ & $0.47_{0.19}$ & $0.59_{0.13}$ & $0.6_{0.14}$ & $0.74_{0.11}$ \\ 

ARC Diff & $-$ & $-$ & $-$ & $-$ & $-$ & $-$ & $0.15_{0.29}$ & $0.28_{0.13}$ & $0.33_{0.31}$ & $0.3_{0.26}$ \\

Abductive NLI & $0.08_{0.26}$ & $\textbf{0.17}_{0.05}$ & $0.16_{0.09}$ & $0.19_{0.19}$ & $0.26_{0.08}$ & $\textbf{0.3}_{0.07}$ & $0.42_{0.13}$ & $0.57_{0.08}$ & $0.61_{0.07}$ & $0.68_{0.07}$ \\ 

Winogrande & $-0.19_{0.11}$ & $-0.03_{0.06}$ & $\textbf{0.0}_{0.17}$ & $-0.11_{0.09}$ & $-0.05_{0.12}$ & $\textbf{0.11}_{0.15}$ & $0.09_{0.14}$ & $0.03_{0.1}$ & $0.14_{0.1}$ & $0.21_{0.14}$ \\ 

CSQA & $0.29_{0.11}$ & $0.28_{0.1}$ & $\textbf{0.31}_{0.07}$ & $0.36_{0.14}$ & $0.37_{0.08}$ & $\textbf{0.39}_{0.09}$ & $0.49_{0.09}$ & $0.69_{0.08}$ & $0.78_{0.04}$ & $0.83_{0.05}$ \\ 

QuaRel & $-$ & $-$ & $-$ & $-$ & $-$ & $-$ & $0.32_{0.26}$ & $0.33_{0.25}$ & $0.39_{0.07}$ & $0.51_{0.1}$ \\ 

QuaRTz & $-$ & $-$ & $-$ & $-$ & $-$ & $-$ & $0.34_{0.19}$ & $0.36_{0.04}$ & $0.34_{0.12}$ & $0.37_{0.08}$ \\ 

% 0.28	0.1	0.3	0.11	0.39	0.13	0.31	0.11	0.35	0.11	0.43	0.14	0.46	0.17	0.58	0.11	0.66	0.08	0.72	0.07	0.74	0.07	0.77	0.07	0.81	0.06	0.84	0.05	0.87	0.04	0.89	0.03
\midrule
\textbf{Average} & $0.28_{0.1}$ & $0.3_{0.11}$ & $\textbf{0.39}_{0.13}$ & $0.31_{0.11}$ & $0.35_{0.11}$ & $\textbf{0.43}_{0.14}$ & $0.46_{0.17}$ & $0.58_{0.11}$ & $0.66_{0.08}$ & $0.72_{0.07}$ \\ 

\bottomrule
\end{tabular}}
\caption{\textbf{Kendall correlation with full evaluation dataset} achieved by various instance selection approaches for different percentage of instances. 
Each cell shows the mean and standard deviation obtained from 5 different runs. $-$ cell indicates 0 selected instances. We show the expanded version of this table in supplementary.}
\label{tab:efficiency_improved}
\end{table*}

%% file: tables/erroneous_examples.tex
\begin{table}[t]
    \small
    \centering
    \resizebox{\linewidth}{!}{
    % \begin{tabular}{@{}p{0.1\linewidth}>{\RaggedRight}p{0.9\linewidth}>{\RaggedRight}@{}}
    \begin{tabular}{p{0.6cm}p{7cm}}
    \toprule
        \textbf{Dataset} &
        \multicolumn{1}{c}{\textbf{Instance}}
        \\
    \midrule

    % \multirow{1}{*}{SNLI}
    SNLI (72\%)
      	& 
    \textbf{Premise}: Trucks racing.	\textbf{Hypothesis}: \underline{Four} trucks are racing against each other in the relay. \newline \textbf{Label:} \textcolor{red}{Entailment}, \textcolor{blue}{Neutral} \\
      
    %   (72\%)
    %   	& 
    %   \textbf{Premise}: Two elderly men having a conversation.	\textbf{Hyp.}: Two elderly \underline{woman} having a conversation with their children. \textcolor{red}{Neutral}, \textcolor{blue}{Contradiction} \\ \\

    % \multirow{1}{*}{CSQA}
      CSQA (50\%)
      	& 
      Why would a band be performing when there are no people nearby?	\textbf{O1}: \textcolor{orange}{record album}, 	\textbf{O2}: play music, 	\textbf{O3}: hold concert, 	\textbf{O4}: blaring,	\textbf{O5}: \textcolor{blue}{practice}  \\ 
    %   (50\%)
    %   	& 
    %   What do audiences clap for?	\textbf{O1}: \textcolor{blue}{cinema}, 	\textbf{O2}: \textcolor{orange}{theatre}, 	\textbf{O3}: \textcolor{blue}{movies}, 	\textbf{O4}: \textcolor{blue}{show},	\textbf{O5}: \textcolor{blue}{hockey game}  \\ \\

    %   \multirow{1}{*}{WG}
    WG (36\%)
     & 
      Maria was able to keep their weight off long term, unlike Felicia, because \_ followed a healthy diet. \newline
      \textbf{O1}: \textcolor{blue}{Maria}, 	\textbf{O2}: \textcolor{red}{Felicia}  \\  
    %   (36\%)
    %   	& 
    %   When Derrick told Christopher about quitting school to provide for their family, \_ started panicking.
    %   \textbf{O1}: \textcolor{red}{Derrick}, 	\textbf{O2}: \textcolor{blue}{Christopher}  \\  \\
    
%   \multirow{1}{*}{aNLI}	
    aNLI (x\%)
   
      & 
      \textbf{O1}: Ella was taking her final exam.	\textbf{O2}: Ella was able to finish her exam on time.	\textbf{H1}: \textcolor{blue}{Ella got to class early and was in no hurry.} \textbf{H2}: \textcolor{red}{Ella broke her pencil.} \\
    %   (x\%)
    %   & 
    %   \textbf{O1}: Cathy was happy that she finally had some time to sew.	\textbf{O2}: Cathy tapped her metal fingertips on the table in frustration.	\textbf{H1}: \textcolor{red}{Cathy put the thimbles on.} \textbf{H2}: \textcolor{blue}{Cathy could not get the thread into the fabric.} \\

    \bottomrule

    \end{tabular}
    }
    \caption{Examples of \textbf{erroneous instances} from SNLI, CSQA, Winogrande, and Abductive NLI. \textcolor{orange} {Orange} (ambiguous) and \textcolor{red}{red} (mislabeled) correspond to the originally annotated answer while \textcolor{blue}{blue} corresponds to the correct/equally probable answer.} 
    \label{tab:erroneous_examples}
\end{table}

%% file: tables/modified_examples.tex
\begin{table}[t]
    \small
    \centering
    \resizebox{\linewidth}{!}{
    % \begin{tabular}{@{}p{0.1\linewidth}>{\RaggedRight}p{0.9\linewidth}>{\RaggedRight}@{}}
    \begin{tabular}{p{4cm}p{4cm}}
    \toprule
        \multicolumn{1}{c}{\textbf{Original Instance}} &
        \multicolumn{1}{c}{\textbf{Modification}}
        \\
    \midrule

    %   \textbf{Premise}: Two dogs play together on the bed.	\textbf{Hyp.}: Two dogs wrestling with a sheet on a bed. \textcolor{red}{E}, \textcolor{blue}{N}\\
    \textbf{P}: A man standing in front of a chalkboard points at a drawing.	\textbf{H}: \textcolor{red}{A kid washes a chalkboard} \textbf{L:} Contradiction
      &
    \textbf{H'}: \textcolor{blue}{A 4 year old male standing in front of a chalkboard points at a drawing.} \newline
    \textbf{Predicted L:} Neutral\\ \\
    
    \textbf{P}: A man is performing tricks with his superbike.	\textbf{H}: \textcolor{red}{A bike is in the garage.} \textbf{L:} Contradiction
      &
    \textbf{H'}: \textcolor{blue}{He is performing stunts on a four wheeler.} \newline
    \textbf{Predicted L:} Neutral\\

    \midrule    
    \textbf{P}: A skateboarder does a trick at a skate park.	\textbf{H}: The skateboarder is performing a heelie kick flip. \textbf{L:} \textcolor{red}{Entailment}
      &
    
    \textbf{L':} \textcolor{blue}{Neutral} \\ \\
    
    \textbf{P}: A little blond girl is running near a little blond boy.	\textbf{H}: A sister and brother are playing in their yard. \textbf{L:} \textcolor{red}{Entailment}
      &
    
    \textbf{L':} \textcolor{blue}{Neutral} \\
    \bottomrule

    \end{tabular}
    }
    \caption{Illustrative \textbf{examples from SNLI dataset modified using our technique}. Top two correspond to trivial instances for which a human modified the hypothesis in a label-preserving manner such that the model's prediction changed. Bottom two correspond to mislabeled instances where the human rectified the label.} 
    \label{tab:modified_examples}
\end{table}

%% file: tables/efficient_eval_supp.tex
\begin{table*}[]
\centering
% \resizebox{\textwidth}{!}
{%
\begin{tabular}{l|llllll}
\toprule
\multicolumn{1}{c}{} &
\multicolumn{1}{c}{\textbf{25\%}} &
\multicolumn{1}{c}{\textbf{30\%}} &
\multicolumn{1}{c}{\textbf{40\%}} &
\multicolumn{1}{c}{\textbf{50\%}} &
\multicolumn{1}{c}{\textbf{60\%}} & 
\multicolumn{1}{c}{\textbf{75\%}} 
\\
\textbf{Dataset} & \textbf{P} & \textbf{P} & \textbf{P} & \textbf{P} & \textbf{P} & \textbf{P}  \\

\midrule
% order - paws qqp, qqp, dnli, paws wiki cola, swag, snli, mnli, adv nli, qnli, sst2, rte, arpc, agnews, emotions, arc, abductive, winogrande, csqa, quarel, quartz, socialiqa 

% snli, paws wiki, adv nli R1, ag news, qnli, mrpc, social iqa, 

SNLI & $0.95_{0.0}$ & $0.95_{0.01}$ & $0.96_{0.01}$ & $0.96_{0.01}$ & $0.96_{0.01}$ & $0.97_{0.01}$ \\ 

PAWS Wiki & $0.98_{0.01}$ & $0.98_{0.02}$ & $0.98_{0.01}$ & $0.98_{0.01}$ & $0.98_{0.01}$ & $0.99_{0.01}$ \\ 

AgNews & $0.93_{0.01}$ & $0.93_{0.02}$ & $0.93_{0.01}$ & $0.96_{0.01}$ & $0.96_{0.01}$ & $0.97_{0.01}$ \\ 

QNLI & $0.92_{0.02}$ & $0.92_{0.03}$ & $0.93_{0.02}$ & $0.96_{0.01}$ & $0.96_{0.01}$ & $0.97_{0.01}$ \\ 

MRPC & $0.67_{0.13}$ & $0.7_{0.11}$ & $0.75_{0.08}$ & $0.84_{0.05}$ & $0.84_{0.03}$ & $0.88_{0.03}$ \\ 

SocialIQA & $0.84_{0.04}$ & $0.87_{0.02}$ & $0.89_{0.02}$ & $0.91_{0.01}$ & $0.93_{0.02}$ & $0.94_{0.03}$ \\ 

QQP & $0.96_{0.01}$ & $0.96_{0.01}$ & $0.96_{0.01}$ & $0.97_{0.01}$ & $0.98_{0.0}$ & $0.99_{0.01}$ \\ 

DNLI & $0.96_{0.02}$ & $0.97_{0.02}$ & $0.97_{0.02}$ & $0.98_{0.01}$ & $0.98_{0.01}$ & $0.98_{0.01}$ \\ 

COLA & $0.8_{0.05}$ & $0.82_{0.07}$ & $0.89_{0.06}$ & $0.91_{0.02}$ & $0.92_{0.04}$ & $0.96_{0.02}$ \\ 

SWAG & $0.97_{0.01}$ & $0.96_{0.01}$ & $0.97_{0.01}$ & $0.98_{0.01}$ & $0.99_{0.0}$ & $0.99_{0.01}$ \\ 

PAWS QQP & $0.89_{0.02}$ & $0.92_{0.02}$ & $0.92_{0.02}$ & $0.93_{0.02}$ & $0.94_{0.01}$ & $0.94_{0.02}$ \\

MNLI & $0.95_{0.01}$ & $0.97_{0.01}$ & $0.97_{0.01}$ & $0.98_{0.0}$ & $0.97_{0.01}$ & $0.98_{0.01}$ \\ 

Adv. NLI R1 & $0.62_{0.06}$ & $0.64_{0.08}$ & $0.67_{0.06}$ & $0.73_{0.06}$ & $0.79_{0.05}$ & $0.84_{0.07}$ \\ 

Adv. NLI R2 & $0.42_{0.08}$ & $0.46_{0.1}$ & $0.54_{0.14}$ & $0.63_{0.05}$ & $0.71_{0.05}$ & $0.77_{0.03}$ \\ 

Adv. NLI R3 & $0.61_{0.05}$ & $0.59_{0.06}$ & $0.66_{0.1}$ & $0.75_{0.06}$ & $0.79_{0.06}$ & $0.85_{0.04}$ \\ 

SST-2 & $0.83_{0.05}$ & $0.86_{0.04}$ & $0.87_{0.02}$ & $0.87_{0.04}$ & $0.91_{0.03}$ & $0.92_{0.01}$ \\ 

ARC Easy & $0.76_{0.07}$ & $0.78_{0.08}$ & $0.84_{0.08}$ & $0.85_{0.05}$ & $0.89_{0.03}$ & $0.94_{0.02}$ \\ 

ARC Diff & $0.41_{0.36}$ & $0.49_{0.32}$ & $0.62_{0.28}$ & $0.59_{0.18}$ & $0.75_{0.1}$ & $0.86_{0.06}$ \\ 

Abductive NLI & $0.72_{0.03}$ & $0.77_{0.03}$ & $0.79_{0.06}$ & $0.82_{0.03}$ & $0.86_{0.02}$ & $0.88_{0.04}$ \\ 

Winogrande & $0.24_{0.13}$ & $0.3_{0.16}$ & $0.39_{0.17}$ & $0.44_{0.16}$ & $0.53_{0.09}$ & $0.63_{0.07}$ \\ 

CSQA & $0.85_{0.04}$ & $0.86_{0.03}$ & $0.89_{0.03}$ & $0.91_{0.02}$ & $0.94_{0.02}$ & $0.95_{0.01}$ \\ 

QuaRel & $0.57_{0.12}$ & $0.58_{0.16}$ & $0.73_{0.1}$ & $0.8_{0.15}$ & $0.79_{0.07}$ & $0.81_{0.07}$ \\ 

QuaRTz & $0.37_{0.12}$ & $0.44_{0.07}$ & $0.51_{0.12}$ & $0.57_{0.09}$ & $0.62_{0.11}$ & $0.64_{0.08}$ \\

\bottomrule
\end{tabular}}
\caption{Kendall correlation with full dataset evaluation achieved by our proposed instance selection approach for different percentage of instances. 
Each cell shows the mean and standard deviation obtained from 5 different runs.}
\label{tab:supp_efficiency_improved}
\end{table*}

%% file: tables/actually_difficult_examples.tex
\begin{table}[t]
    \small
    \centering
    \resizebox{\linewidth}{!}{
    % \begin{tabular}{@{}p{0.1\linewidth}>{\RaggedRight}p{0.9\linewidth}>{\RaggedRight}@{}}
    \begin{tabular}{p{8cm}}
    \toprule
        \multicolumn{1}{c}{\textbf{Difficult Instance}}
        \\
    \midrule

    \textbf{Premise}: Dog standing with 1 foot up in a large field.	\textbf{Hyp.}: The dog is standing on one leg. \textbf{Label:} Contradiction. \\ \\
    
    \textbf{Premise}: A salt-and-pepper-haired man with beard and glasses wearing black sits on the grass.	\textbf{Hyp.}: An elderly bearded man sitting on the grass. \textbf{Label:} Entailment. \\ \\
    
    \textbf{Premise}: A man is standing in front of a building holding heart shaped balloons and a woman is crossing the street.	\textbf{Hyp.}: Someone is holding something heavy outside. \textbf{Label:} Contradiction. \\ \\
    
    \textbf{Premise}: A group of people plays a game on the floor of a living room while a TV plays in the background.	\textbf{Hyp.}: A group of friends are playing the xbox while other friends wait for their turn. \textbf{Label:} Contradiction. \\

    \bottomrule

    \end{tabular}
    }
    \caption{Illustrative examples of instances that receive high difficulty score but are not erroneous. Such instances are difficult even for humans as they require reasoning ability.} 
    \label{tab:supp_actually_difficult_examples}
\end{table}

%% file: tables/supp_erroneous.tex
\begin{table}[t]
    \small
    \centering
    \resizebox{\linewidth}{!}{
    % \begin{tabular}{@{}p{0.1\linewidth}>{\RaggedRight}p{0.9\linewidth}>{\RaggedRight}@{}}
    \begin{tabular}{p{0.6cm}p{7cm}}
    \toprule
        \textbf{Dataset} &
        \multicolumn{1}{c}{\textbf{Instance}}
        \\
    \midrule

    \multirow{1}{*}{SNLI}
    % SNLI (72\%)
      	& 
    \textbf{Premise}: Trucks racing.	\textbf{Hyp.}: \underline{Four} trucks are racing against each other in the relay. \textcolor{red}{Entailment}, \textcolor{blue}{Neutral} \\
      
      (72\%)
      	& 
      \textbf{Premise}: Two elderly men having a conversation.	\textbf{Hyp.}: Two elderly \underline{woman} having a conversation with their children. \textcolor{red}{Neutral}, \textcolor{blue}{Contradiction} \\ \\

    \multirow{1}{*}{CSQA}
    %   CSQA (50\%)
      	& 
      Why would a band be performing when there are no people nearby?	\textbf{O1}: \textcolor{orange}{record album}, 	\textbf{O2}: play music, 	\textbf{O3}: hold concert, 	\textbf{O4}: blaring,	\textbf{O5}: \textcolor{blue}{practice}  \\ 
      (50\%)
      	& 
      What do audiences clap for?	\textbf{O1}: \textcolor{blue}{cinema}, 	\textbf{O2}: \textcolor{orange}{theatre}, 	\textbf{O3}: \textcolor{blue}{movies}, 	\textbf{O4}: \textcolor{blue}{show},	\textbf{O5}: \textcolor{blue}{hockey game}  \\ \\

      \multirow{1}{*}{WG}
    % WG (36\%)
     & 
      Maria was able to keep their weight off long term, unlike Felicia, because \_ followed a healthy diet.
      \textbf{O1}: \textcolor{blue}{Maria}, 	\textbf{O2}: \textcolor{red}{Felicia}  \\  
      (36\%)
      	& 
      When Derrick told Christopher about quitting school to provide for their family, \_ started panicking.
      \textbf{O1}: \textcolor{red}{Derrick}, 	\textbf{O2}: \textcolor{blue}{Christopher}  \\  \\
    
  \multirow{1}{*}{aNLI}	
    % aNLI (x\%)
   
      & 
      \textbf{O1}: Ella was taking her final exam.	\textbf{O2}: Ella was able to finish her exam on time.	\textbf{H1}: \textcolor{blue}{Ella got to class early and was in no hurry.} \textbf{H2}: \textcolor{red}{Ella broke her pencil.} \\
      (x\%)
      & 
      \textbf{O1}: Cathy was happy that she finally had some time to sew.	\textbf{O2}: Cathy tapped her metal fingertips on the table in frustration.	\textbf{H1}: \textcolor{red}{Cathy put the thimbles on.} \textbf{H2}: \textcolor{blue}{Cathy could not get the thread into the fabric.} \\

    \bottomrule

    \end{tabular}
    }
    \caption{Illustrative examples of erroneous instances in SNLI, CSQA, Winogrande, and Abductive NLI. \textcolor{orange} {Orange} (ambiguous) and \textcolor{red}{red} (mislabeled) indicate  the originally annotated answer while \textcolor{blue}{blue} indicates the True/equally probable answer. } 
    \label{tab:supp_erroneous_examples}
\end{table}

%% file: tables/trivial_examples.tex
\begin{table}[t]
    \small
    \centering
    \resizebox{\linewidth}{!}{
    % \begin{tabular}{@{}p{0.1\linewidth}>{\RaggedRight}p{0.9\linewidth}>{\RaggedRight}@{}}
    \begin{tabular}{p{0.6cm}p{7cm}}
    \toprule
        \textbf{Dataset} &
        \multicolumn{1}{c}{\textbf{Instance}}
        \\
    \midrule

    \multirow{1}{*}{SNLI} 
      	& 
    \textbf{Premise}: A woman playing with her cats while taking pictures.	\textbf{Hyp.}: A woman is playing with her dolls. \textcolor{blue}{Contradiction} \\
      
    \multirow{1}{*}{CSQA}
  
      	& 
      What will a person going for a jog likely be wearing?	\textbf{O1}: grope, 	\textbf{O2}: acknowledgment, 	\textbf{O3}: \textcolor{blue}{comfortable clothes}, 	\textbf{O4}: ipod,	\textbf{O5}: passionate kisses  \\ 
      \multirow{1}{*}{WG}
     & 
      Katrina did not value the antique pictures as much as Lindsey because \_ was a history buff.
      \textbf{O1}: Katrina, 	\textbf{O2}: \textcolor{blue}{Lindsey}  \\  
      	
   \multirow{1}{*}{aNLI}	
   
      & 
      \textbf{O1}: I bought a house with an ugly yard.	\textbf{O2}: He carved the rock into a lion head and kept it.	\textbf{H1}: \textcolor{blue}{There was a large rock in the yard.} \textbf{H2}: I decided to tear the whole notebook up. \\
     
    \bottomrule

    \end{tabular}
    }
    \caption{Illustrative examples of trivial instances in SNLI, CSQA, Winogrande, and Abductive NLI. Text in \textcolor{blue}{blue} corresponds to the ground-truth answer.} 
    \label{tab:trivial_examples}
\end{table}

%% file: acl_latex.bbl
\begin{thebibliography}{51}
\expandafter\ifx\csname natexlab\endcsname\relax\def\natexlab#1{#1}\fi

\bibitem[{Baker and Kim(2004)}]{baker2004item}
Frank~B Baker and Seock-Ho Kim. 2004.
\newblock \emph{Item response theory: Parameter estimation techniques}.
\newblock CRC Press.

\bibitem[{Bhagavatula et~al.(2020)Bhagavatula, Bras, Malaviya, Sakaguchi,
  Holtzman, Rashkin, Downey, tau Yih, and Choi}]{bhagavatula2020abductive}
Chandra Bhagavatula, Ronan~Le Bras, Chaitanya Malaviya, Keisuke Sakaguchi, Ari
  Holtzman, Hannah Rashkin, Doug Downey, Wen tau Yih, and Yejin Choi. 2020.
\newblock \href {https://openreview.net/forum?id=Byg1v1HKDB} {Abductive
  commonsense reasoning}.
\newblock In \emph{International Conference on Learning Representations}.

\bibitem[{Bowman et~al.(2015)Bowman, Angeli, Potts, and
  Manning}]{bowman-etal-2015-large}
Samuel~R. Bowman, Gabor Angeli, Christopher Potts, and Christopher~D. Manning.
  2015.
\newblock \href {https://doi.org/10.18653/v1/D15-1075} {A large annotated
  corpus for learning natural language inference}.
\newblock In \emph{Proceedings of the 2015 Conference on Empirical Methods in
  Natural Language Processing}, pages 632--642, Lisbon, Portugal. Association
  for Computational Linguistics.

\bibitem[{Clark et~al.(2020)Clark, Luong, Le, and Manning}]{clark2020electra}
Kevin Clark, Minh-Thang Luong, Quoc~V. Le, and Christopher~D. Manning. 2020.
\newblock \href {https://openreview.net/pdf?id=r1xMH1BtvB} {{ELECTRA}:
  Pre-training text encoders as discriminators rather than generators}.
\newblock In \emph{ICLR}.

\bibitem[{Clark et~al.(2018)Clark, Cowhey, Etzioni, Khot, Sabharwal, Schoenick,
  and Tafjord}]{clark2018think}
Peter Clark, Isaac Cowhey, Oren Etzioni, Tushar Khot, Ashish Sabharwal, Carissa
  Schoenick, and Oyvind Tafjord. 2018.
\newblock Think you have solved question answering? try arc, the ai2 reasoning
  challenge.
\newblock \emph{arXiv preprint arXiv:1803.05457}.

\bibitem[{Devlin et~al.(2019)Devlin, Chang, Lee, and
  Toutanova}]{devlin-etal-2019-bert}
Jacob Devlin, Ming-Wei Chang, Kenton Lee, and Kristina Toutanova. 2019.
\newblock \href {https://doi.org/10.18653/v1/N19-1423} {{BERT}: Pre-training of
  deep bidirectional transformers for language understanding}.
\newblock In \emph{Proceedings of the 2019 Conference of the North {A}merican
  Chapter of the Association for Computational Linguistics: Human Language
  Technologies, Volume 1 (Long and Short Papers)}, pages 4171--4186,
  Minneapolis, Minnesota. Association for Computational Linguistics.

\bibitem[{Dolan and Brockett(2005)}]{dolan2005automatically}
William~B Dolan and Chris Brockett. 2005.
\newblock Automatically constructing a corpus of sentential paraphrases.
\newblock In \emph{Proceedings of the Third International Workshop on
  Paraphrasing (IWP2005)}.

\bibitem[{Gururangan et~al.(2018)Gururangan, Swayamdipta, Levy, Schwartz,
  Bowman, and Smith}]{gururangan-etal-2018-annotation}
Suchin Gururangan, Swabha Swayamdipta, Omer Levy, Roy Schwartz, Samuel Bowman,
  and Noah~A. Smith. 2018.
\newblock \href {https://doi.org/10.18653/v1/N18-2017} {Annotation artifacts in
  natural language inference data}.
\newblock In \emph{Proceedings of the 2018 Conference of the North {A}merican
  Chapter of the Association for Computational Linguistics: Human Language
  Technologies, Volume 2 (Short Papers)}, pages 107--112, New Orleans,
  Louisiana. Association for Computational Linguistics.

\bibitem[{Hendrycks and Dietterich(2019)}]{hendrycks2018benchmarking}
Dan Hendrycks and Thomas Dietterich. 2019.
\newblock \href {https://openreview.net/forum?id=HJz6tiCqYm} {Benchmarking
  neural network robustness to common corruptions and perturbations}.
\newblock In \emph{International Conference on Learning Representations}.

\bibitem[{Hendrycks et~al.(2020)Hendrycks, Liu, Wallace, Dziedzic, Krishnan,
  and Song}]{hendrycks-etal-2020-pretrained}
Dan Hendrycks, Xiaoyuan Liu, Eric Wallace, Adam Dziedzic, Rishabh Krishnan, and
  Dawn Song. 2020.
\newblock \href {https://doi.org/10.18653/v1/2020.acl-main.244} {Pretrained
  transformers improve out-of-distribution robustness}.
\newblock In \emph{Proceedings of the 58th Annual Meeting of the Association
  for Computational Linguistics}, pages 2744--2751, Online. Association for
  Computational Linguistics.

\bibitem[{Iandola et~al.(2020)Iandola, Shaw, Krishna, and
  Keutzer}]{iandola-etal-2020-squeezebert}
Forrest Iandola, Albert Shaw, Ravi Krishna, and Kurt Keutzer. 2020.
\newblock \href {https://doi.org/10.18653/v1/2020.sustainlp-1.17}
  {{S}queeze{BERT}: What can computer vision teach {NLP} about efficient neural
  networks?}
\newblock In \emph{Proceedings of SustaiNLP: Workshop on Simple and Efficient
  Natural Language Processing}, pages 124--135, Online. Association for
  Computational Linguistics.

\bibitem[{Iyer et~al.(2017)Iyer, Dandekar, and Csernai}]{iyer2017first}
Shankar Iyer, Nikhil Dandekar, and Korn{\'e}l Csernai. 2017.
\newblock First quora dataset release: Question pairs.
\newblock \emph{data. quora. com}.

\bibitem[{Jiang et~al.(2020)Jiang, Yu, Zhou, Chen, Feng, and
  Yan}]{NEURIPS2020_96da2f59}
Zi-Hang Jiang, Weihao Yu, Daquan Zhou, Yunpeng Chen, Jiashi Feng, and Shuicheng
  Yan. 2020.
\newblock \href
  {https://proceedings.neurips.cc/paper/2020/file/96da2f590cd7246bbde0051047b0d6f7-Paper.pdf}
  {Convbert: Improving bert with span-based dynamic convolution}.
\newblock In \emph{Advances in Neural Information Processing Systems},
  volume~33, pages 12837--12848. Curran Associates, Inc.

\bibitem[{Kendall(1938)}]{kendall1938new}
Maurice~G Kendall. 1938.
\newblock A new measure of rank correlation.
\newblock \emph{Biometrika}, 30(1/2):81--93.

\bibitem[{Kiela et~al.(2021)Kiela, Bartolo, Nie, Kaushik, Geiger, Wu, Vidgen,
  Prasad, Singh, Ringshia, Ma, Thrush, Riedel, Waseem, Stenetorp, Jia, Bansal,
  Potts, and Williams}]{kiela-etal-2021-dynabench}
Douwe Kiela, Max Bartolo, Yixin Nie, Divyansh Kaushik, Atticus Geiger,
  Zhengxuan Wu, Bertie Vidgen, Grusha Prasad, Amanpreet Singh, Pratik Ringshia,
  Zhiyi Ma, Tristan Thrush, Sebastian Riedel, Zeerak Waseem, Pontus Stenetorp,
  Robin Jia, Mohit Bansal, Christopher Potts, and Adina Williams. 2021.
\newblock \href {https://doi.org/10.18653/v1/2021.naacl-main.324} {Dynabench:
  Rethinking benchmarking in {NLP}}.
\newblock In \emph{Proceedings of the 2021 Conference of the North American
  Chapter of the Association for Computational Linguistics: Human Language
  Technologies}, pages 4110--4124, Online. Association for Computational
  Linguistics.

\bibitem[{Lalor et~al.(2018)Lalor, Wu, Munkhdalai, and
  Yu}]{lalor-etal-2018-understanding}
John~P. Lalor, Hao Wu, Tsendsuren Munkhdalai, and Hong Yu. 2018.
\newblock \href {https://doi.org/10.18653/v1/D18-1500} {Understanding deep
  learning performance through an examination of test set difficulty: A
  psychometric case study}.
\newblock In \emph{Proceedings of the 2018 Conference on Empirical Methods in
  Natural Language Processing}, pages 4711--4716, Brussels, Belgium.
  Association for Computational Linguistics.

\bibitem[{Lalor et~al.(2016)Lalor, Wu, and Yu}]{lalor-etal-2016-building}
John~P. Lalor, Hao Wu, and Hong Yu. 2016.
\newblock \href {https://doi.org/10.18653/v1/D16-1062} {Building an evaluation
  scale using item response theory}.
\newblock In \emph{Proceedings of the 2016 Conference on Empirical Methods in
  Natural Language Processing}, pages 648--657, Austin, Texas. Association for
  Computational Linguistics.

\bibitem[{Lan et~al.(2020)Lan, Chen, Goodman, Gimpel, Sharma, and
  Soricut}]{Lan2020ALBERT:}
Zhenzhong Lan, Mingda Chen, Sebastian Goodman, Kevin Gimpel, Piyush Sharma, and
  Radu Soricut. 2020.
\newblock \href {https://openreview.net/forum?id=H1eA7AEtvS} {Albert: A lite
  bert for self-supervised learning of language representations}.
\newblock In \emph{International Conference on Learning Representations}.

\bibitem[{Liu et~al.(2019)Liu, Ott, Goyal, Du, Joshi, Chen, Levy, Lewis,
  Zettlemoyer, and Stoyanov}]{Liu2019RoBERTaAR}
Yinhan Liu, Myle Ott, Naman Goyal, Jingfei Du, Mandar Joshi, Danqi Chen, Omer
  Levy, M.~Lewis, Luke Zettlemoyer, and Veselin Stoyanov. 2019.
\newblock Roberta: A robustly optimized bert pretraining approach.
\newblock \emph{ArXiv}, abs/1907.11692.

\bibitem[{McCoy et~al.(2020)McCoy, Min, and Linzen}]{mccoy-etal-2020-berts}
R.~Thomas McCoy, Junghyun Min, and Tal Linzen. 2020.
\newblock \href {https://doi.org/10.18653/v1/2020.blackboxnlp-1.21} {{BERT}s of
  a feather do not generalize together: Large variability in generalization
  across models with similar test set performance}.
\newblock In \emph{Proceedings of the Third BlackboxNLP Workshop on Analyzing
  and Interpreting Neural Networks for NLP}, pages 217--227, Online.
  Association for Computational Linguistics.

\bibitem[{McCoy et~al.(2019)McCoy, Pavlick, and Linzen}]{mccoy-etal-2019-right}
Tom McCoy, Ellie Pavlick, and Tal Linzen. 2019.
\newblock \href {https://doi.org/10.18653/v1/P19-1334} {Right for the wrong
  reasons: Diagnosing syntactic heuristics in natural language inference}.
\newblock In \emph{Proceedings of the 57th Annual Meeting of the Association
  for Computational Linguistics}, pages 3428--3448, Florence, Italy.
  Association for Computational Linguistics.

\bibitem[{Mishra and Arunkumar(2021)}]{mishra2021robust}
Swaroop Mishra and Anjana Arunkumar. 2021.
\newblock How robust are model rankings: A leaderboard customization approach
  for equitable evaluation.
\newblock In \emph{Proceedings of the AAAI Conference on Artificial
  Intelligence}, volume~35, pages 13561--13569.

\bibitem[{Mishra et~al.(2020)Mishra, Arunkumar, Sachdeva, Bryan, and
  Baral}]{mishra2020dqi}
Swaroop Mishra, Anjana Arunkumar, Bhavdeep Sachdeva, Chris Bryan, and Chitta
  Baral. 2020.
\newblock Dqi: Measuring data quality in nlp.
\newblock \emph{arXiv preprint arXiv:2005.00816}.

\bibitem[{Mishra and Sachdeva(2020)}]{mishra-sachdeva-2020-need}
Swaroop Mishra and Bhavdeep~Singh Sachdeva. 2020.
\newblock \href {https://doi.org/10.18653/v1/2020.sustainlp-1.23} {Do we need
  to create big datasets to learn a task?}
\newblock In \emph{Proceedings of SustaiNLP: Workshop on Simple and Efficient
  Natural Language Processing}, pages 169--173, Online. Association for
  Computational Linguistics.

\bibitem[{Nie et~al.(2020)Nie, Williams, Dinan, Bansal, Weston, and
  Kiela}]{nie-etal-2020-adversarial}
Yixin Nie, Adina Williams, Emily Dinan, Mohit Bansal, Jason Weston, and Douwe
  Kiela. 2020.
\newblock \href {https://doi.org/10.18653/v1/2020.acl-main.441} {Adversarial
  {NLI}: A new benchmark for natural language understanding}.
\newblock In \emph{Proceedings of the 58th Annual Meeting of the Association
  for Computational Linguistics}, pages 4885--4901, Online. Association for
  Computational Linguistics.

\bibitem[{Rodriguez et~al.(2021)Rodriguez, Barrow, Hoyle, Lalor, Jia, and
  Boyd-Graber}]{rodriguez-etal-2021-evaluation}
Pedro Rodriguez, Joe Barrow, Alexander~Miserlis Hoyle, John~P. Lalor, Robin
  Jia, and Jordan Boyd-Graber. 2021.
\newblock \href {https://doi.org/10.18653/v1/2021.acl-long.346} {Evaluation
  examples are not equally informative: How should that change {NLP}
  leaderboards?}
\newblock In \emph{Proceedings of the 59th Annual Meeting of the Association
  for Computational Linguistics and the 11th International Joint Conference on
  Natural Language Processing (Volume 1: Long Papers)}, pages 4486--4503,
  Online. Association for Computational Linguistics.

\bibitem[{Sagawa et~al.(2020)Sagawa, Raghunathan, Koh, and
  Liang}]{sagawa2020investigation}
Shiori Sagawa, Aditi Raghunathan, Pang~Wei Koh, and Percy Liang. 2020.
\newblock An investigation of why overparameterization exacerbates spurious
  correlations.
\newblock In \emph{International Conference on Machine Learning}, pages
  8346--8356. PMLR.

\bibitem[{Sakaguchi et~al.(2020)Sakaguchi, Bras, Bhagavatula, and
  Choi}]{Sakaguchi2020WINOGRANDEAA}
Keisuke Sakaguchi, Ronan~Le Bras, Chandra Bhagavatula, and Yejin Choi. 2020.
\newblock Winogrande: An adversarial winograd schema challenge at scale.
\newblock In \emph{AAAI}.

\bibitem[{Sanh et~al.(2019)Sanh, Debut, Chaumond, and
  Wolf}]{Sanh2019DistilBERTAD}
Victor Sanh, Lysandre Debut, Julien Chaumond, and Thomas Wolf. 2019.
\newblock Distilbert, a distilled version of bert: smaller, faster, cheaper and
  lighter.
\newblock \emph{ArXiv}, abs/1910.01108.

\bibitem[{Sap et~al.(2019)Sap, Rashkin, Chen, Bras, and Choi}]{Sap2019SocialIC}
Maarten Sap, Hannah Rashkin, Derek Chen, Ronan~Le Bras, and Yejin Choi. 2019.
\newblock Social iqa: Commonsense reasoning about social interactions.
\newblock In \emph{EMNLP 2019}.

\bibitem[{Socher et~al.(2013)Socher, Perelygin, Wu, Chuang, Manning, Ng, and
  Potts}]{socher-etal-2013-recursive}
Richard Socher, Alex Perelygin, Jean Wu, Jason Chuang, Christopher~D. Manning,
  Andrew Ng, and Christopher Potts. 2013.
\newblock \href {https://aclanthology.org/D13-1170} {Recursive deep models for
  semantic compositionality over a sentiment treebank}.
\newblock In \emph{Proceedings of the 2013 Conference on Empirical Methods in
  Natural Language Processing}, pages 1631--1642, Seattle, Washington, USA.
  Association for Computational Linguistics.

\bibitem[{Swayamdipta et~al.(2020)Swayamdipta, Schwartz, Lourie, Wang,
  Hajishirzi, Smith, and Choi}]{swayamdipta-etal-2020-dataset}
Swabha Swayamdipta, Roy Schwartz, Nicholas Lourie, Yizhong Wang, Hannaneh
  Hajishirzi, Noah~A. Smith, and Yejin Choi. 2020.
\newblock \href {https://doi.org/10.18653/v1/2020.emnlp-main.746} {Dataset
  cartography: Mapping and diagnosing datasets with training dynamics}.
\newblock In \emph{Proceedings of the 2020 Conference on Empirical Methods in
  Natural Language Processing (EMNLP)}, pages 9275--9293, Online. Association
  for Computational Linguistics.

\bibitem[{Tafjord et~al.(2019{\natexlab{a}})Tafjord, Clark, Gardner, Yih, and
  Sabharwal}]{tafjord2019quarel}
Oyvind Tafjord, Peter Clark, Matt Gardner, Wen-tau Yih, and Ashish Sabharwal.
  2019{\natexlab{a}}.
\newblock Quarel: A dataset and models for answering questions about
  qualitative relationships.
\newblock In \emph{Proceedings of the AAAI Conference on Artificial
  Intelligence}, volume~33, pages 7063--7071.

\bibitem[{Tafjord et~al.(2019{\natexlab{b}})Tafjord, Gardner, Lin, and
  Clark}]{tafjord-etal-2019-quartz}
Oyvind Tafjord, Matt Gardner, Kevin Lin, and Peter Clark. 2019{\natexlab{b}}.
\newblock \href {https://doi.org/10.18653/v1/D19-1608} {{Q}ua{RT}z: An
  open-domain dataset of qualitative relationship questions}.
\newblock In \emph{Proceedings of the 2019 Conference on Empirical Methods in
  Natural Language Processing and the 9th International Joint Conference on
  Natural Language Processing (EMNLP-IJCNLP)}, pages 5941--5946, Hong Kong,
  China. Association for Computational Linguistics.

\bibitem[{Talmor et~al.(2019)Talmor, Herzig, Lourie, and
  Berant}]{talmor-etal-2019-commonsenseqa}
Alon Talmor, Jonathan Herzig, Nicholas Lourie, and Jonathan Berant. 2019.
\newblock \href {https://doi.org/10.18653/v1/N19-1421} {{C}ommonsense{QA}: A
  question answering challenge targeting commonsense knowledge}.
\newblock In \emph{Proceedings of the 2019 Conference of the North {A}merican
  Chapter of the Association for Computational Linguistics: Human Language
  Technologies, Volume 1 (Long and Short Papers)}, pages 4149--4158,
  Minneapolis, Minnesota. Association for Computational Linguistics.

\bibitem[{Tan et~al.(2019)Tan, Shen, Huang, and
  Courville}]{Tan2019InvestigatingBI}
Shawn Tan, Yikang Shen, Chin-Wei Huang, and Aaron~C. Courville. 2019.
\newblock Investigating biases in textual entailment datasets.
\newblock \emph{ArXiv}, abs/1906.09635.

\bibitem[{Vania et~al.(2021)Vania, Htut, Huang, Mungra, Pang, Phang, Liu, Cho,
  and Bowman}]{vania-etal-2021-comparing}
Clara Vania, Phu~Mon Htut, William Huang, Dhara Mungra, Richard~Yuanzhe Pang,
  Jason Phang, Haokun Liu, Kyunghyun Cho, and Samuel~R. Bowman. 2021.
\newblock \href {https://doi.org/10.18653/v1/2021.acl-long.92} {Comparing test
  sets with item response theory}.
\newblock In \emph{Proceedings of the 59th Annual Meeting of the Association
  for Computational Linguistics and the 11th International Joint Conference on
  Natural Language Processing (Volume 1: Long Papers)}, pages 1141--1158,
  Online. Association for Computational Linguistics.

\bibitem[{Wang et~al.(2019)Wang, Pruksachatkun, Nangia, Singh, Michael, Hill,
  Levy, and Bowman}]{NEURIPS2019_4496bf24}
Alex Wang, Yada Pruksachatkun, Nikita Nangia, Amanpreet Singh, Julian Michael,
  Felix Hill, Omer Levy, and Samuel Bowman. 2019.
\newblock \href
  {https://proceedings.neurips.cc/paper/2019/file/4496bf24afe7fab6f046bf4923da8de6-Paper.pdf}
  {Superglue: A stickier benchmark for general-purpose language understanding
  systems}.
\newblock In \emph{Advances in Neural Information Processing Systems},
  volume~32. Curran Associates, Inc.

\bibitem[{Wang et~al.(2018)Wang, Singh, Michael, Hill, Levy, and
  Bowman}]{wang-etal-2018-glue}
Alex Wang, Amanpreet Singh, Julian Michael, Felix Hill, Omer Levy, and Samuel
  Bowman. 2018.
\newblock \href {https://doi.org/10.18653/v1/W18-5446} {{GLUE}: A multi-task
  benchmark and analysis platform for natural language understanding}.
\newblock In \emph{Proceedings of the 2018 {EMNLP} Workshop {B}lackbox{NLP}:
  Analyzing and Interpreting Neural Networks for {NLP}}, pages 353--355,
  Brussels, Belgium. Association for Computational Linguistics.

\bibitem[{Warstadt et~al.(2019)Warstadt, Singh, and
  Bowman}]{warstadt-etal-2019-neural}
Alex Warstadt, Amanpreet Singh, and Samuel~R. Bowman. 2019.
\newblock \href {https://doi.org/10.1162/tacl_a_00290} {Neural network
  acceptability judgments}.
\newblock \emph{Transactions of the Association for Computational Linguistics},
  7:625--641.

\bibitem[{Weiss(1982)}]{weiss1982improving}
David~J Weiss. 1982.
\newblock Improving measurement quality and efficiency with adaptive testing.
\newblock \emph{Applied psychological measurement}, 6(4):473--492.

\bibitem[{Welleck et~al.(2019)Welleck, Weston, Szlam, and
  Cho}]{welleck-etal-2019-dialogue}
Sean Welleck, Jason Weston, Arthur Szlam, and Kyunghyun Cho. 2019.
\newblock \href {https://doi.org/10.18653/v1/P19-1363} {Dialogue natural
  language inference}.
\newblock In \emph{Proceedings of the 57th Annual Meeting of the Association
  for Computational Linguistics}, pages 3731--3741, Florence, Italy.
  Association for Computational Linguistics.

\bibitem[{Williams et~al.(2018)Williams, Nangia, and
  Bowman}]{williams-etal-2018-broad}
Adina Williams, Nikita Nangia, and Samuel Bowman. 2018.
\newblock \href {https://doi.org/10.18653/v1/N18-1101} {A broad-coverage
  challenge corpus for sentence understanding through inference}.
\newblock In \emph{Proceedings of the 2018 Conference of the North {A}merican
  Chapter of the Association for Computational Linguistics: Human Language
  Technologies, Volume 1 (Long Papers)}, pages 1112--1122, New Orleans,
  Louisiana. Association for Computational Linguistics.

\bibitem[{Xu et~al.(2020)Xu, Zhang, Mao, Wang, Xie, and
  Zhang}]{xu-etal-2020-curriculum}
Benfeng Xu, Licheng Zhang, Zhendong Mao, Quan Wang, Hongtao Xie, and Yongdong
  Zhang. 2020.
\newblock \href {https://doi.org/10.18653/v1/2020.acl-main.542} {Curriculum
  learning for natural language understanding}.
\newblock In \emph{Proceedings of the 58th Annual Meeting of the Association
  for Computational Linguistics}, pages 6095--6104, Online. Association for
  Computational Linguistics.

\bibitem[{Yang et~al.(2019{\natexlab{a}})Yang, Dai, Yang, Carbonell,
  Salakhutdinov, and Le}]{Yang2019XLNetGA}
Zhilin Yang, Zihang Dai, Yiming Yang, J.~Carbonell, R.~Salakhutdinov, and
  Quoc~V. Le. 2019{\natexlab{a}}.
\newblock Xlnet: Generalized autoregressive pretraining for language
  understanding.
\newblock In \emph{NeurIPS}.

\bibitem[{Yang et~al.(2019{\natexlab{b}})Yang, Dai, Yang, Carbonell,
  Salakhutdinov, and Le}]{NEURIPS2019_dc6a7e65}
Zhilin Yang, Zihang Dai, Yiming Yang, Jaime Carbonell, Russ~R Salakhutdinov,
  and Quoc~V Le. 2019{\natexlab{b}}.
\newblock \href
  {https://proceedings.neurips.cc/paper/2019/file/dc6a7e655d7e5840e66733e9ee67cc69-Paper.pdf}
  {Xlnet: Generalized autoregressive pretraining for language understanding}.
\newblock In \emph{Advances in Neural Information Processing Systems},
  volume~32. Curran Associates, Inc.

\bibitem[{Zellers et~al.(2018)Zellers, Bisk, Schwartz, and
  Choi}]{zellers2018swagaf}
Rowan Zellers, Yonatan Bisk, Roy Schwartz, and Yejin Choi. 2018.
\newblock Swag: A large-scale adversarial dataset for grounded commonsense
  inference.
\newblock In \emph{Proceedings of the 2018 Conference on Empirical Methods in
  Natural Language Processing (EMNLP)}.

\bibitem[{Zhang et~al.(2015)Zhang, Zhao, and LeCun}]{NIPS2015_250cf8b5}
Xiang Zhang, Junbo Zhao, and Yann LeCun. 2015.
\newblock \href
  {https://proceedings.neurips.cc/paper/2015/file/250cf8b51c773f3f8dc8b4be867a9a02-Paper.pdf}
  {Character-level convolutional networks for text classification}.
\newblock In \emph{Advances in Neural Information Processing Systems},
  volume~28. Curran Associates, Inc.

\bibitem[{Zhang et~al.(2019)Zhang, Baldridge, and He}]{zhang-etal-2019-paws}
Yuan Zhang, Jason Baldridge, and Luheng He. 2019.
\newblock \href {https://doi.org/10.18653/v1/N19-1131} {{PAWS}: Paraphrase
  adversaries from word scrambling}.
\newblock In \emph{Proceedings of the 2019 Conference of the North {A}merican
  Chapter of the Association for Computational Linguistics: Human Language
  Technologies, Volume 1 (Long and Short Papers)}, pages 1298--1308,
  Minneapolis, Minnesota. Association for Computational Linguistics.

\bibitem[{Zhong et~al.(2021)Zhong, Ghosh, Klein, and
  Steinhardt}]{zhong-etal-2021-larger}
Ruiqi Zhong, Dhruba Ghosh, Dan Klein, and Jacob Steinhardt. 2021.
\newblock \href {https://doi.org/10.18653/v1/2021.findings-acl.334} {Are larger
  pretrained language models uniformly better? comparing performance at the
  instance level}.
\newblock In \emph{Findings of the Association for Computational Linguistics:
  ACL-IJCNLP 2021}, pages 3813--3827, Online. Association for Computational
  Linguistics.

\bibitem[{Zhou et~al.(2020)Zhou, Nie, Tan, and Bansal}]{zhou-etal-2020-curse}
Xiang Zhou, Yixin Nie, Hao Tan, and Mohit Bansal. 2020.
\newblock \href {https://doi.org/10.18653/v1/2020.emnlp-main.659} {The curse of
  performance instability in analysis datasets: Consequences, source, and
  suggestions}.
\newblock In \emph{Proceedings of the 2020 Conference on Empirical Methods in
  Natural Language Processing (EMNLP)}, pages 8215--8228, Online. Association
  for Computational Linguistics.

\end{thebibliography}
